\pdfoutput=1

\documentclass[11pt]{article}

\usepackage[final]{acl} 

\usepackage{times}
\usepackage{latexsym}

\usepackage{blindtext}
\usepackage{hyperref}
\usepackage{enumitem}

\usepackage[T1]{fontenc}
\usepackage{listings}

\usepackage[utf8]{inputenc}
\usepackage{multirow}

\usepackage{booktabs}
\usepackage{adjustbox}

\usepackage{graphicx}
\usepackage{subcaption}

\usepackage{microtype}

\usepackage{inconsolata}

\title{Effects of diversity incentives on sample diversity and downstream model performance in LLM-based text augmentation}

\author{Jan Cegin$^{\spadesuit}$$^\dagger$, Branislav Pecher$^{\spadesuit}$$^\dagger$, Jakub Simko$^\dagger$, Ivan Srba$^\dagger$ Maria Bielikova$^\dagger$,
{\bf Peter Brusilovsky}$^\ddagger$ \\
  $^{\spadesuit}$ Faculty of Information Technology, Brno University of Technology, Brno, Czechia \\
  $^\dagger$ Kempelen Institute of Intelligent Technologies, Bratislava, Slovakia\\
  \texttt{\{jan.cegin, branislav.pecher, jakub.simko, ivan.srba, maria.bielikova\}}@kinit.sk \\
  $^\ddagger$ University of Pittsburgh, Pittsburgh, USA \\
  \texttt{peterb@pitt.edu }}

\begin{document}
	\maketitle
	\begin{abstract}
		The latest generative large language models (LLMs) have found their application in data augmentation tasks, where small numbers of text samples are LLM-paraphrased and then used to fine-tune downstream models. However, more research is needed to assess how different prompts, seed data selection strategies, filtering methods, or model settings affect the quality of paraphrased data (and downstream models). In this study, we investigate three text diversity incentive methods well established in crowdsourcing: \emph{taboo} words, \emph{hints} by previous outlier solutions, and \emph{chaining} on previous outlier solutions. Using these incentive methods as part of instructions to LLMs augmenting text datasets, we measure their effects on generated texts' lexical diversity and downstream model performance. We compare the effects over 5 different LLMs, 6 datasets and 2 downstream models. We show that diversity is most increased by \emph{taboo} words, but downstream model performance is highest with \emph{hints}.
	\end{abstract}
	
\section{Introduction}
    The emergence of large language models (LLMs) such as GPT-4, LLaMA, etc., has sparked interest in using them to augment textual datasets~\cite{ubani2023zeroshotdataaug, dai2023auggpt, piedboeuf-langlais-2023-chatgpt}. In these scenarios, the number of samples is expanded by paraphrasing existing ones through LLM prompting. The created paraphrases are then added to the original dataset and used for downstream model training. Such methods have been explored for various domains such as sentiment classification~\cite{piedboeuf-langlais-2023-chatgpt, ubani2023zeroshotdataaug}, news classification~\cite{piedboeuf-langlais-2023-chatgpt} and health symptoms classifications~\cite{dai2023auggpt}. However, investigation of the effect of various prompts, specific instructions, and selection of seed data inspired by crowd in the text augmentation process when using LLMs is lacking.

    \begin{table}[t!]
    \centering
    \small
    \setlength\tabcolsep{3pt}
    \begin{tabular}{@{}lcccccc@{}}
    \toprule
   \textsc{Method}$\rightarrow$ & \multicolumn{2}{c}{\textsc{Taboo}} & \multicolumn{2}{c}{\textsc{Chaining}} & \multicolumn{2}{c}{\textsc{Hints}}\\
   Dataset$\downarrow$  & BERT & Mistral & BERT & Mistral & BERT & Mistral\\
     \midrule
    20News & $\textcolor{black}{0}/\textcolor{blue}{1}$ & $\textcolor{black}{0}/\textcolor{blue}{2}$  & $\textcolor{orange}{1}/\textcolor{blue}{1}$ & $\textcolor{black}{0}/\textcolor{black}{0}$ & $\textcolor{black}{0}/\textcolor{blue}{1}$ & $\textcolor{black}{0}/\textcolor{blue}{4}$ \\
    AG News & $\textcolor{black}{0}/\textcolor{black}{0}$ & $\textcolor{black}{0}/\textcolor{blue}{2}$  & $\textcolor{black}{0}/\textcolor{black}{0}$ & $\textcolor{black}{0}/\textcolor{blue}{2}$ & $\textcolor{black}{0}/\textcolor{black}{0}$ & $\textcolor{black}{0}/\textcolor{blue}{3}$ \\
    ATIS & $\textcolor{orange}{2}/\textcolor{black}{0}$ & $\textcolor{orange}{1}/\textcolor{black}{0}$  & $\textcolor{black}{0}/\textcolor{black}{0}$ & $\textcolor{black}{0}/\textcolor{black}{0}$ & $\textcolor{black}{0}/\textcolor{black}{0}$ & $\textcolor{black}{0}/\textcolor{blue}{1}$ \\
    FB & $\textcolor{orange}{1}/\textcolor{black}{0}$ & $\textcolor{orange}{1}/\textcolor{blue}{1}$  & $\textcolor{black}{0}/\textcolor{black}{0}$ & $\textcolor{black}{0}/\textcolor{black}{0}$ & $\textcolor{black}{0}/\textcolor{blue}{1}$ & $\textcolor{black}{0}/\textcolor{blue}{2}$ \\
    SST-5 & $\textcolor{black}{0}/\textcolor{black}{0}$ & $\textcolor{black}{0}/\textcolor{blue}{2}$  & $\textcolor{black}{0}/\textcolor{black}{0}$ & $\textcolor{black}{0}/\textcolor{blue}{1}$ & $\textcolor{black}{0}/\textcolor{black}{0}$ & $\textcolor{black}{0}/\textcolor{blue}{2}$ \\
    Yelp & $\textcolor{black}{0}/\textcolor{black}{0}$ & $\textcolor{black}{0}/\textcolor{blue}{2}$  & $\textcolor{black}{0}/\textcolor{black}{0}$ & $\textcolor{black}{0}/\textcolor{black}{0}$ & $\textcolor{black}{0}/\textcolor{black}{0}$ & $\textcolor{black}{0}/\textcolor{blue}{2}$ \\
    \bottomrule
    \end{tabular}
    \caption{Number of \textcolor{blue}{overperforming} or \textcolor{orange}{underperforming} cases of downstream models (BERT, Mistral) fine-tuned on LLM-generated data. We observe a \textbf{varying performance when diversity incentive methods are used} during the training set generation. Only the \emph{hints} incentive method appears to frequently \textcolor{blue}{significantly outperform the \emph{baseline}} (= no incentives), indicating its potential usefulness in LLM augmentation. \emph{Taboo} and \emph{chaining} methods achieve mixed results, sometimes even \textcolor{orange}{dropping below the \emph{baseline}}. The effects appear more frequently with fine-tuned Mistral than BERT. The rows denote 6 datasets used in the experiments. We used 5 different LLMs to generate the training sets for each dataset-method-model combination.}
    \label{tab:res_aggregated}
    \end{table}

    Crowdsourcing is an established practice for collecting training or validation examples for a variety of NLP tasks. Scenarios of data collection using human workers can be similar to those of data augmentation: workers create paraphrases on existing sentences chosen from a dataset. The aim of such data collection is to increase the \textit{data diversity} and subsequent performance of classifiers trained on the data~\cite{Larson2019, larson-etal-2020-iterative}. To increase the diversity, various methods are used in crowdsourcing to guide workers. These include \textit{taboo} words~\cite{larson-etal-2020-iterative} - where most significant words from the collected data are identified and listed in the worker instructions to be avoided during paraphrasing, \textit{chaining}~\cite{Cox2021, Larson2019} - where outliers in the previous paraphrases are identified and used as seed sentences in the next round of data collection, and \textit{hints} where previous outlier paraphrases are used as examples in the instructions. The \textit{hints}~\cite{Cox2021, Yaghoub-Zadeh-Fard2020} method itself is similar to LLM in-context learning, where examples are included in the instructions for the model to achieve better performance. All of these \textit{diversity incentive methods} report increased diversity of paraphrases and some also report increased performance of the classifiers trained on the so-collected data.

    This work is inspired by the parallels between crowdsourcing and LLM prompting and by the performance of \textit{diversity incentive methods} on the diversity of paraphrases and the performance of models trained on them. We investigate the effects of the three \textit{diversity incentive} methods (originating in crowdsourcing) on data augmentation using LLMs. The baseline, taken from a previous study~\cite{cegin-etal-2023-chatgpt}, is a simple prompting for paraphrases. Measuring paraphrase diversity and downstream performance of classification models, we assess whether the diversity incentives (added to the base prompt) improve LLM outputs similarly as in crowdsourcing scenarios. To our knowledge, this is the first work to investigate the effects of \textit{diversity incentive methods} on LLMs.

    In this paper, we answer the following research questions:
	\begin{description}[labelwidth = 24pt, leftmargin = !]
		\item[RQ1:] \emph{Does the usage of diversity incentive methods on LLMs yield more diverse paraphrases? (compared to base prompting)}
		\item[RQ2:] \emph{Do classifiers achieve better performance if trained on data augmented using diversity incentive methods on LLMs? (compared to base prompting)}
	\end{description}

    To answer these questions~\footnote{Data and code at: \url{https://github.com/kinit-sk/LLM-div-incts}}, we have conducted a data augmentation experiment using 5 different LLMs on 6 different datasets in the tasks of sentiment (movie and app reviews), news, and intent (flight and voice assistant commands) classification. In this experiment, we repeatedly collect LLM paraphrases using different diversity incentive methods. Then, we compare the \emph{lexical diversity} of the collected data and the \emph{performance of downstream classifiers}. Additionally, we also conduct an \textit{ablation study}, where we modify the diversity incentive methods with random data to validate, that the inputs used by these methods (e.g., most influential taboo words, outlier paraphrases) contribute to the method's performance and a combination of the best performing methods for lexical diversity and model performance. In total, we collected 253,500 paraphrases.
    
    The most prominent findings are the following: 1) We do not observe statistically significant improvements in lexical diversity of the generated datasets, but only minor improvements using the \textit{taboo} method,
    2) The \textit{hints} method increases the performance of classification models trained on such data compared to the baseline, while also reducing standard deviation and thus increasing the stability of results,
    3) The \textit{chaining} method and \textit{taboo} method both do not  significantly affect the performance of classification models trained on such data compared to the baseline.

\section{Related work: Crowdsourcing and LLM-based augmentation}

    \subsection{Crowdsourcing diverse paraphrases}

    Crowdsourcing of paraphrases is an established method to collect data for dataset building and augmentation in NLP~\cite{Larson2019, larson-etal-2020-iterative, Yaghoub-Zadeh-Fard2020, Bohlen2018, Cox2021}. In this process, a worker is asked to paraphrase a seed sentence to create new variants~\cite{Bohlen2018, larson-etal-2020-iterative}. To increase the diversity of paraphrases, various instruction variants are used, building on the assumption (shown by~\cite{larson-etal-2020-iterative, joshi-he-2022-investigation, wang2022toward}), that performance of downstream models correlates with training set diversity.

    The \textit{hints} method~\cite{Cox2021, Yaghoub-Zadeh-Fard2020} guides workers towards a variety of possible solutions by showing them examples of the most distinct paraphrases previously created by other workers. A variation of this method displays word-clouds of recommended words to be used. \textit{Hints} have been used for the data collection of user utterances for task-oriented chatbots and to collect diverse motivational messages.   

    The \textit{taboo} method~\cite{larson-etal-2020-iterative} instructs workers to avoid specific words when paraphrasing. These ``taboo words'' are drawn from previously collected paraphrases as most influential using a linear SVM. Taboo words have been used in the collection of data for intent classification.

    The \textit{chaining} method~\cite{Cox2021, Larson2019} identifies outliers or most distinct paraphrases within the already collected data and uses them as seed sentences. It is applied in variations for data collection of intent utterances and of motivational messages.

    In crowdsourcing, all three methods show increases in the paraphrase diversity and model performance, compared to base prompting.

    \subsection{Data augmentation via LLMs}

    LLMs such as GPT-2~\cite{radford2019language} or BART~\cite{lewis-etal-2020-bart} have previously been used to create paraphrases. Additional extensions used style transfer to create paraphrases of a certain linguistic style~\cite{Krishna2020}, syntax control of the generated paraphrases~\cite{goyal-durrett-2020-neural, chen-etal-2020-semantically}, multi-lingual paraphrases in a zero-shot setting~\cite{thompson-post-2020-paraphrase} and LLM finetuning using Low-Rank Adaption for specific domain paraphrase collection~\cite{chowdhury2022novelty}. Recent studies used GPT-3.5 and GPT-4 as data augmentation techniques that were compared with previous state-of-art NLP augmentation techniques~\cite{piedboeuf-langlais-2023-chatgpt, ubani2023zeroshotdataaug}. Two studies report better performance in using LLMs as data augmenters than using previous state-of-art techniques in both paraphrasing existing texts~\cite{dai2023auggpt} and in a zero-shot setup of generating new texts using specific prompts~\cite{ubani2023zeroshotdataaug}. Another study reports mixed results, when GPT-3.5 is compared with previous state-of-the-art techniques~\cite{piedboeuf-langlais-2023-chatgpt}. Regardless of mixed results, GPT-like models have already been used as augmenters in domains of automated scoring~\cite{fang2023using} and low-resource language generation~\cite{ghosh-etal-2023-dale}. However, LLMs can also produce repeating outputs of lower quality~\cite{cegin-etal-2023-chatgpt, cox-2023-prompting}.

    To our best knowledge, there is no study which investigates if and how diversity incentives (established in crowdsourcing), can be used in LLMs-based paraphrasing. We hypothesize, that use of diversity incentives can prevent the known drawback of LLMs to generate highly similar repetitive content (a challenge that was addressed in crowdsourcing by diversity incentives).

-\section{Data collection and evaluation methodology}\label{sec:data_coll}

      \begin{figure*}[ht]
        \centering
        \includegraphics[width=16cm]{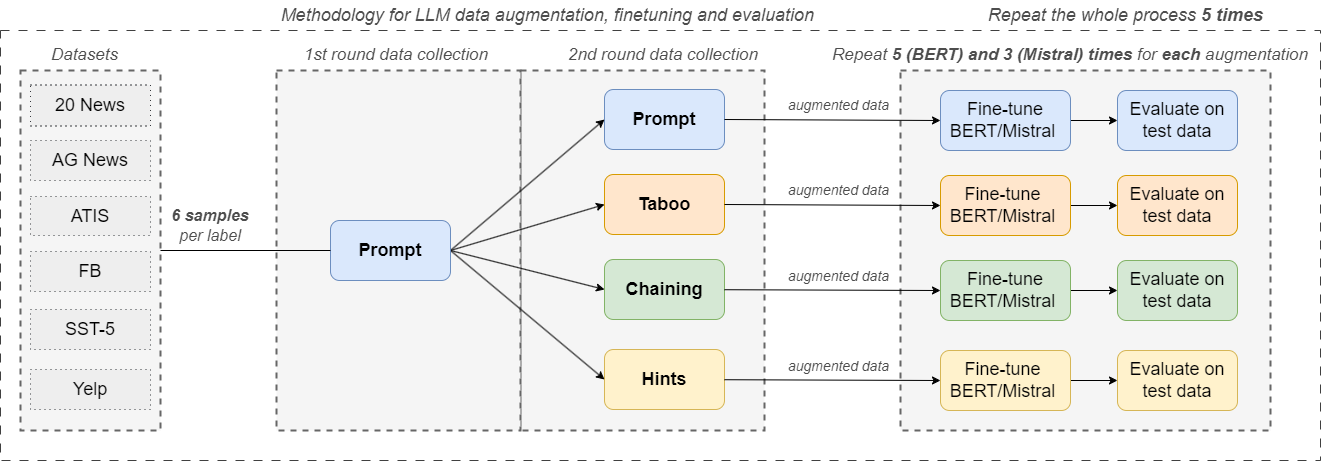}
        \caption{Overview of our methodology. For each dataset, we randomly sample 6 samples per label that are used as seed sentences for LLM data augmentation. There, we collect data in in 2 rounds - 1st only using the \textit{prompt} method and then in parallel for \textit{prompt} method and 3 different diversity incentive methods. These are added together to form the datasets. BERT-large or Mistral classifier is fine-tuned 5 or 3 times respectively on each of the collected data and then evaluated. We repeat the entire process 5 times.}
        \label{fig:dataset_build}
    \end{figure*}

    We collected paraphrases for all combinations of the following: 5 different LLMs, 6 datasets, and 3 diversity incentive methods + 1 base prompting. For each combination, 5 collection iterations were performed: in each 6 random seed sentences per label were drawn from a dataset. For each prompt fired, 5 paraphrases were collected. This totalled in 142,500 collected paraphrases when aggregated all together across datasets and LLMs. For the ablation study and combination of best methods in Section~\ref{sec:comb_methods} we collected an additional 111,000 paraphrases in total.
    

    As the diversity incentive methods need some previously collected data to determine their cues (hints, seeds or taboo words), each iteration consisted of 2 rounds: first we collected data using only the basic prompt and in the second round, we collected data using the given diversity incentive method (or base prompt method). Thus, the resulting datasets for each method consist of seed data and data collected from both rounds. The entire data collection process is visualized in Figure~\ref{fig:dataset_build}.

    After the paraphrases were collected, we evaluated them in several steps. First, we manually checked the \emph{validity} of a subset (50\%) of the collected data (i.e., is the created sample a true paraphrase retaining the label?). Second, we computed the \emph{diversity} of the collected data, comparing the mean vocabulary size (no. unique words) and mean number of unique 3-grams for each diversity incentive method (refers to RQ1).

    Third, we evaluated the performance of models trained on the created paraphrases (refers to RQ2). For each combination of LLM, dataset and method, we finetuned BERT-large 5 times and Mistral-7b-v0.1 3 times (the dataset also determined the classification task to which a model was finetuned). We evaluated the accuracy of trained model on the full test set for that given dataset specifically and on a subset of the test set for Mistral to save computational resources following previous works~\cite{chang-jia-2023-data, koksal-etal-2023-meal, li-qiu-2023-finding, gao-etal-2021-making}, as the inference time is long and costly. Details of the finetuning process can be found in Appendix~\ref{sec:appendix_bert_finetune} and~\ref{sec:appendix_mistral_finetune}.

    \subsection{Prompt design}\label{sec:div_incen_method}

    As our base prompt, we adopted the instruction design from a previous LLM-paraphrasing study~\cite{cegin-etal-2023-chatgpt}. There, the prompt plainly instructs to \textit{``Paraphrase this text or sentence 5 times:''}, which is followed by the seed sentence.

    For the \textit{taboo} method we take the implementation from~\cite{larson-etal-2020-iterative} that uses a linear SVM trained on bag-of-words representation in a one-vs-many setting to identify the 3 most significant words that are then used in the instructions. We run the computation to get taboo words on the first round of collected data that were collected using the \textit{prompt} method. We also filter out named entities using NLTK to not include them as taboo words. The prompt used in this method is taken from~\cite{cegin-etal-2023-chatgpt}.

    For the \textit{chaining} method we use an outlier detection method from~\cite{Larson2019} where we first compute per label a mean embedding vector from the collected samples from the first round. Then, using Euclidean distance, we find the collected samples that are the furthest away from the mean vector to be used as seed sentences in the second round of data collection. The prompt used in this method is the same as in the \textit{prompt} method.

    The \textit{hints} method is similar to the previous approach with \textit{chaining} where we find outliers in the collected data the same way. Here the data are only included in the prompt itself as examples listed with the given seed sentence. The listed examples are always only those that have been created from the given seed to be paraphrased. The prompt is the same as in the \textit{prompt} method with added delimiter section listing the 3 different hints for the seed sentence.
    
    Templates and examples of the prompts can be found in Appendix~\ref{sec:appendix_example-incentives}.
    
    \subsection{LLMs used as generators}
    We used 5 different LLMs as data augmenters - 2 open source LLMs, LLaMA-2 and 2 closed LLMs. 
    We chose open LLMs based on their different performance and size on the OpenLLM leaderboard. We used the instruction finetuned versions of the LLMs available at HuggingFace. Namely, for LLaMA-2~\cite{touvron2023LLaMA2} we use LLaMA-2-70B-instruct, for Platypus~\cite{platypus2023} we use Platypus-70B-instruct and for Mistral~\cite{jiang2023mistral} we use Mistral-7B-instruct. We collected the data on a custom private infrastructure with 16 core CPU, 64 GB RAM and 4xA100 GPUs. As for the closed LLMs, we used 2 of the most widely used: GPT3.5 denoted as ChatGPT (\textit{gpt-3.5-turbo-1106} version) and GPT4 (\textit{gpt-4-0613 version}).

    \subsection{Datasets used}
    We used 6 different datasets for our data collection experiments from the domains of news, intent and sentiment classification. We specifically focused on multi-class English datasets as the \textit{diversity incentive} methods were employed in crowdsourcing processes that used multi-class English datasets. We used  the \textit{20 news}~\cite{LANG1995331} and \textit{AG news}~\cite{zhang2015character} datasets for news classification, \textit{FB}~\cite{schuster-etal-2019-cross-lingual} and \textit{ATIS}~\cite{hemphill-etal-1990-atis} datasets for intent classification and \textit{SST-5}~\cite{socher-etal-2013-recursive} and \textit{Yelp}~\cite{zhang2015character} datasets for sentiment classification. We did not use all of the labels in our experiments for the news and intent classification datasets, but randomly select a subset of them. More details can be found in Appendix~\ref{sec:appendix_dataset_details}.

    \subsection{Ablation study setup}\label{sec:ablt_setup}
    To investigate if the diversity incetive methods actually influence the diversity of the collected data and performance of classifiers trained on such data we conduct an ablation study. Here, we repeat the data collection process for the open-source LLMs (Mistral, Platypus) and LLaMA-2 using modified versions of each of the diversity incentive method to investigate whether the particular setup of the methods themselves as they have been used in crowdsourcing literature influences the results.

    For the \textit{taboo} method, instead of using the most significant words from the previously generated paraphrases we used 3 random words from the generated paraphrases. For the \textit{chaining} method and \textit{hints} method, instead of using the outliers as the next seed sentences or as a hints, we used any previously generated paraphrase that was randomly chosen as seed or hint respectively.

    \section{Paraphrase validity and diversity}\label{sec:diversity_validit}

    \subsection{Validity of paraphrases}
    
    Before evaluating validity of paraphrases, we filtered for malformed phrases, empty phrases or duplicated phrases as per~\cite{cegin-etal-2023-chatgpt}. As we collect only 5 samples per one seed sentence, we have detected no duplicated phrases. There were some malformed phrases generated by all LLMs with the exception of ChatGPT, but their number was generally low. The number of collected samples per dataset can be found in Appendix~\ref{sec:appendix_no_dataset_col}. The highest amount of mangled or empty paraphrases were detected in GPT-4 responses, mostly when using the \textit{chaining} method, where the number of invalid paraphrases was approx. 5\%. For Mistral, LLaMA-2 and Platypus we detected around 1\% of mangled paraphrases. We found no impact of diversity incentive on the number of mangled or empty paraphrases for these LLMs. The detected mangled or empty paraphrases were removed and not included in the next stages.

    Second, for each dataset and LLM combination, we sampled 50\% of the collected data to be manually validated, i.e. we checked whether the resulting paraphrases are semantically equivalent to the seed sentences and their labels. Details are in Appendix~\ref{sec:appendix_para_valid}. Among diversity incentive methods, we detected no invalid utterance, in line with the findings of~\cite{cegin-etal-2023-chatgpt}.

    \subsection{Lexical diversity of paraphrases}

    Next, we investigated the effect of diversity incentive methods on the lexical diversity of the collected datasets. We focused on the number of collected unique words (vocabulary) and the number of collected unique 3-grams for each dataset. As we repeated the data collection process 5 times for each dataset and LLM combination, we report the mean numbers of collected unique words and 3-grams. We visualize our findings in Appendix~\ref{sec:appendix_lex_vis_res}.
    
    In nearly all cases except for one (ChatGPT for the \textit{AG News} dataset) the \textit{taboo} method yielded a higher-than-baseline number of unique words and 3-grams. The \textit{hints} and \textit{chaining} methods yielded only occasional increases in lexical diversity, with fluctuating results of increased and decreased lexical diversity across LLMs and datasets. However, the resulting increases in lexical diversity were not statistically significant, as we investigated using the Wilcoxon signed-rank test (\textit{p=0.05}).  
    
    In more details, the \textit{taboo} method increased mean no. unique words 30/30 cases and no. unique 3-grams 29/30 cases. The \textit{chaining} method had better diversity than the baseline in 9/30 cases for unique ngrams and in 4/30 cases the diversity was similar. It achieved better diversity in 10/30 cases for unique words and similar in 9/30 cases. The \textit{hints} method yielded similar results, achieving better no. of unique ngrams than the baseline 10/30 cases and similar in 5/30 cases, while achieving better no. unique words in 9/30 cases and 8/30 cases it was similar to the baseline. The relative increase in lexical diversity ranges from approx. 2\% (Yelp, SST-5 datasets) to 10 \% (ATIS, FB datasets) for both no. unique words and 3-grams. 
    
    In summary, even thought the \textit{taboo} method increases the lexical diversity in nearly all of the cases, the increase is not statistically significant. This contrasts with the crowdsourcing literature. It indicates that the LLMs are using lexically rich vocabulary already with the base prompting, hence the low benefit of diversity incentive methods.

    \subsection{Ablation study results}

    Here, we compared the number of collected n-grams and words between the ablated and non-ablated diversity incentive methods. We label methods as of similar performance if the difference in the number of collected 3-grams or words is less than 10 and we also perform statistical tests. We report the difference between non-ablated and ablated methods in Figures~\ref{fig:ablation_res_words_div} and~\ref{fig:ablation_res_ngrams_div}.

    The non-ablated \textit{taboo} method has betters results in both words (19/30 cases better, 8/30 similar, 3/30 worse) and n-grams (22/30 cases better, 7/30 similar, 1/30 worse) collected than its ablated counterpart. This indicates that the use of the most significant words helps LLMs generate more diverse data in most cases. In contrast, the non-ablated \textit{chaining} and \textit{hints} methods yield better diversity in only 8/30 cases for number of unique words and even less so for the number of unique 3-grams. In more than half of the cases the lexical diversity decreased. This indicates that the usage of outliers as seed sentences or as examples is not desirable when targeting higher lexical diversity.

    We answer the \textit{RQ1: Does the usage of diversity incentive methods on LLMs yield more diverse paraphrases?} as follows: the usage of the \textit{taboo} method increases the lexical diversity of collected data when compared to both the baseline method and the ablated version of the method itself. Other two methods however affect the diversity of collected paraphrases only randomly. These changes are, however, not statistically significant, indicating that the LLMs use rich lexical vocabulary even without the diversity incentives themselves.

    \section{Finetuning models on data collected via diversity incentive methods}\label{sec:model_perf}

    To investigate whether the \textit{diversity incentive} methods improve the performance of downstream models, we finetuned BERT-large 5 times and Mistral 3 times for each LLM-dataset combination. Additionally, as we work with limited data, which was found to cause large variance and instability in finetuning results~\cite{mosbach2020stability, mosbach-etal-2023-shot, pecher2023effects, chang-jia-2023-data}, we sampled data 5 times. This resulted in 25 finetuned classifiers for BERT (5 data collection rounds and 5 finetunings for each of those data collection rounds) and 15 for Mistral that we evaluate per dataset-LLM combination. The full details about hyperparameters and the finetuning setup of BERT and Mistral classifier can be found in Appendices~\ref{sec:appendix_bert_finetune} and~\ref{sec:appendix_mistral_finetune} respectively. We report the accuracy of the finetuned models on the test split of each dataset and focus on 2 main attributes: \emph{mean accuracy} and \emph{stability} of performance (by measuring standard deviation of accuracy).
    Additionally, we also conducted Mann-Whitney-U tests \textit{(p=0.05)} between the baseline \textit{prompt} method and other diversity incentive methods. We are interested in consistent, better performance of a diversity incentive method over the \textit{prompt} baseline across LLMs and datasets, as fluctuating performance could be an indicator of random effects. See summary in Table~\ref{tab:res_aggregated} and full results in the Appendix~\ref{sec:appendix_full_res}.

    \subsection{Impact of diversity incentives on model performance}

    In terms of mean achieved accuracy from all diversity incentive methods while finetuning BERT, the \textit{hints} method achieved best performance across all LLM and dataset combinations by consistently outperforming or achieving similar mean value as the baseline \textit{prompt} method in 28 out of 30 LLM and dataset combinations - 20 cases of better mean performance and 9 cases of similar (difference less than 0.1\%). However, only 3 out of the 19 (15.79\%) increases were statistically significant. Finetuning of Mistral yielded stronger results as the \textit{hints} method achieved better performance 25/30 times, 4/30 times the performance was similar and once worse. Out of the 25 times the \textit{hints} method performed better, 14 times (56\%) it was statistically significant. We speculate that this might be due to better capabilities of the model to use the augmented data. In terms of LLMs used for data augmentation, the statistically significant increases were achieved in 3/6 cases for Platypus and Mistral, in 4/6 cases for LLaMA2 and GPT-4.
    The relative increase in mean performance ranged from 0.6\% to 2.5\% better performance than the baseline for BERT and 1\% to 11\% for Mistral.

    The \textit{taboo} method did significantly worse than the baseline for BERT in 3/30 cases, and only once better. On the other hand, the decrease on Mistral happened in 2/30 cases, while 9/30 times there was a significant increases in performance. The \textit{taboo} method achieved better results on Mistral, similar to the \textit{hints} method. The \textit{chaining} method did not perform better or worse in most cases, yielding a very similar mean performance in most cases for both BERT and Mistral.

    In terms of performance stability, BERT finetuned on data collected via the \textit{hints} method achieved better stability of performance (standard deviation relative difference less than 5\%) in 22/30 cases and similar stability of performance in 5 cases. For Mistral, better stability was achieved in 26/30 cases, with 2 cases of similar and 2 cases of worse stability (on the FB dataset). The relative increase of stability over baseline \textit{prompt} method is from approx. 5\% to 35\% for BERT and from 10\% to 66\% for Mistral.  
    
    The \textit{taboo} method achieves better stability for 14/30 cases, 9/30 cases it is worse than the baseline and 7/30 cases the stability is similar for BERT. For Mistral the results are similar: 15/30 cases the stability is better, 11/30 cases it is worse and 4/30 cases it is similar to the baseline. The \textit{chaining} method achieves better stability of performance only half of the time for BERT and 18/30 cases for Mistral, with 8 cases of worse stability.
    
    In nearly all cases the \textit{hints} method achieves higher mean performance than the baseline \textit{prompt} method and on average achieves higher stability of performance as seen by decreased standard deviation and increased minimum value of finetuned models for both BERT and Mistral. The increases in mean performance and stability are more significant for Mistral than BERT, being statistically significant in 14/26 cases of better performance. The \textit{taboo} method more often than not increases the performance over the baseline and achieves lower stability, but only does so around half of the time, which could indicate random chance at play. Models finetuned on data collected using the \textit{taboo} method can also underperform significantly. The \textit{chaining} method performs similar to the \textit{taboo} method, with fluctuating results in both stability and mean performance.

    \subsection{Ablation study results}

    Similar to the Section~\ref{sec:diversity_validit}, we evaluate the diversity incentive methods also terms of an ablation study conducted via details from Section~\ref{sec:ablt_setup} to investigate whether the setup of the methods themselves contributes to their performance. We visualize our findings in Appendix~\ref{sec:appendix_ablt_res_all}.

    The non-ablated \textit{hints} method has in 29/30 cases better mean performance that the ablated version for BERT and in 27/30 cases for Mistral, with statistically significant results in 8/30 cases for BERT and 4/30 for Mistral. This might indicate that the usage of outliers as hints for the LLMs tends to increase the quality of collected data in data augmentation scenarios when compared to hints chosen randomly as in the ablated method.
    
    The non-ablated \textit{taboo} method achieves better mean performance in 8/18 cases (all of them statistically significant) for BERT and 12/30 cases for Mistral (6 cases of statistical significance). However, the ablated version of \textit{taboo} method was better than the non-ablated version in 4/30 cases significantly. This implies that the use of most significant words as taboo instructions for the LLMs has no significant effect in data augmentation. The non-ablated version of \textit{chaining} method achieves better mean performance in 9/30 cases (4 cases of statistical significance) for BERT and in 12/30 for Mistral (3 cases of statistical significance). For Mistral, in equally 3 cases the results were statistically worse. This implies, similar to the \textit{taboo} method, that the usage of previous outliers as seed sentences has no significant effect on LLMs in a data augmentation scenario when compared to the usage of random previous paraphrases as seed sentences.

    We answer the \textit{RQ2: Do classifiers achieve better performance if trained on data augmented using diversity incentive methods on LLMs?} as follows: only models finetuned on data collected via the \textit{hints} method achieve better stability and mean performance than those trained on data collected via the baseline \textit{prompt} method. The \textit{hints} method also achieves better mean performance and stability of performance when compared to its ablated version. The data collected via the \textit{taboo} and \textit{chaining} methods have random influence on the performance of finetuned models. These results indicate that the usage of outliers as hints for LLMs in a data augmentation scenarios is beneficial, while other methods have no advantage over the baseline of using only prompt instructions.

    \section{Combining diversity incentives}\label{sec:comb_methods}

    As the \textit{taboo} method achieved best results in lexical diversity in and the \textit{hints} method achieved best results in model performance, as follow-up, we decided to combine these two methods to see if we can achieve an improvement. We have performed the data collection and finetuning process in the same way as described in Section~\ref{sec:data_coll}.

    In terms of lexical diversity, the method itself does not have any statistical significance on the results, although the mean number of unique words is higher than the baseline in 18/30 cases and the number of unique n-grams is higher in 16/30 cases. However, in some of the remaining cases a considerable (more than 5\%) drop can be observed. In terms of model performance, the \textit{combined} method statistically significantly decreased the model performance over baseline in 5/30 cases with no increases for BERT and increases performance in 4/30 cases for Mistral. Additionally, it always performed worse as either the \textit{hints} or \textit{taboo} method.

    In summary, the combination of \textit{hints} and \textit{taboo} method into one method grants little to no advantage over either of the methods in both lexical diversity and model performance. We hypothesize that this might be due to the more complicated instructions to the LLM when collecting the data. A decoupling of the methods in a chain of tasks could potentially improve this approach in the future. 

\section{Discussion}
    Given the results of our experiments, we note these following observations: First, contrary to the performance of diversity incentive methods observed by related work in crowdsourcing settings (better lexical diversity of paraphrases and better performance of downstream models), \textbf{not all of the methods show improvement of the lexical diversity when used with LLMs}. The worst performing method is the \textit{chaining} method, where recent works already pointed out that LLMs create progressively worse paraphrases when using their own outputs as seed sentences repeatedly~\cite{tripto2023ship}. However, \textbf{none of the changes in lexical diversity are of statistical significance}.

    Second, the best performing method for data augmentation is the \textit{hints} method, which is similar to in-context learning where demonstrations of samples are provided to the LLM as part of the prompt. This might be the reason why this method works so well, as the own paraphrases of the LLM guide it to better output, similar to in-context learning.

    Third, we observe that, contrary to some previous works~\cite{larson-etal-2020-iterative, joshi-he-2022-investigation}, \textbf{the lexical diversity of the paraphrases does not correlate with performance of models trained on them}. Even though the data collected using the \textit{taboo} method yield highest lexical diversity, models trained on such data do not achieve consistently better performance against baseline.

    Fourth,\textbf{the increase in mean performance and stability seems to be small, but in relative terms (compared to the baseline method) it seems to be significant}, as the increase of mean performance can range from 0.6\% to 2.5\% increase over baseline for BERT and 1\% to 11\% increase for Mistral. For stability, the increases are even more significant: for BERT the range is between 5\% to 35\% increase over baseline and for Mistral from 10\% to 66\%.
    
    Fifth, diversity incentives require additional computations (for significant words and outlier paraphrases) and also require larger LLM context (e.g., hints use additional paraphrases in instructions of the model), meaning higher costs. \textbf{ As such, the increased computation costs may not warrant the use of diversity incentives}. 

    Sixth, \textbf{the combination of the best method for lexical diversity (\textit{taboo}) and best method for model performance (\textit{hints}) did not yield the increases in both lexical diversity and model performance, but performed rather poorly}. We hypothesize that this might be due to the increased context length for the LLM with additional instructions that are hard to perform in one single action.

    \textbf{The promising results using the \textit{hints} method opens possibilities for investigations of in-context learning for text generation in LLMs}, as the quality of such generated data using hints seems to be better than without them. This is in line with the recent results~\cite{cox-2023-prompting} that indicate that the usage of previous examples in instructions for LLMs leads to better generated data.
	
\section{Conclusion} 

    In this work, we investigated the effects of different diversity incentive methods used in crowdsourcing on the lexical diversity of LLM-augmented textual datasets and performance of classification models trained on such data. We compared 3 of such methods with a baseline of using only prompts asking the LLM to paraphrase a given seed. We experimented with 5 LLMs on 6 datasets. Our results indicate that the \textit{taboo} method increases lexical diversity of the collected data, but that this change is not of statistical significance and affects performance only randomly. The \textit{hints} method affects lexical diversity randomly, but increases the performance of classification models (both in stability of and mean performance) that were trained on data collected using this method. The \textit{chaining} method does not improve lexical diversity or model performance of classification models trained on data collected using this method. The combination of \textit{hints} method and \textit{taboo} method does not significantly increase the lexical diversity or model performance. A common downside of diversity incentive methods is the increase of inference costs. Also, there is still some randomness present when using these methods, as even the best performing methods do not increase lexical diversity or performance of models in all cases.

    The notable relative increase in stability of performance and mean performance of models trained on data collected using the \textit{hints} method indicates that LLMs can produce data of better quality using this method when aiming for downstream task classifier performance.

\section*{Limitations}
    We note several limitations to our work. 
    
    First, we did not explore the usability of the diversity incentive methods for languages other than English or for multi-lingual language models.

    Second, we did not use different types of prompts in our experiments and followed those used in previous studies~\cite{cegin-etal-2023-chatgpt, larson-etal-2020-iterative}. Different prompts could have effects on the quality of LLMs, but would radically increase the size of this study, and as such we decided to leave this for future work.

    Third, we evaluated the Mistral finetuned models on only a subset of the test data to save computational resources as some datasets had large test data sets similar to other works~\cite{chang-jia-2023-data, li-qiu-2023-finding, gao-etal-2021-making, koksal-etal-2023-meal}. We did, however, use different splits for each finetuning to mitigate the impact of sample bias.

    Fourth, the worse results of the \textit{taboo} method might be due to the fact that the method sometimes uses unrelated words. Taboo words are determined per label which may yield words with little relevance to the seed sentence. This limitations stems from our replication of the method from crowdsouring, where no such filters were described in the original work of~\cite{larson-etal-2020-iterative}.

    Fifth, we have collected data from only 2 open-source LLMs as well as from LLaMA-2, but we believe that the inclusion of different LLMs and the consistent improvement of \textit{hints} method on model performance across data collected from various LLMs does not threaten our findings.

    Sixth, we only used 2 classifiers for finetuning, namely BERT-large and Mistral-7B-v0.1. However, we believe that repeated data collection rounds and multiple finetunings on the collected data for a variety of datasets mitigates this drawback and as such it does not threaten our findings.

    Seventh, we collected 5 samples per seed sentence and used 6 seed sentences per label from the datasets we used, which resulted in (relatively to the original datasets) smaller datasets (ranging from approx. 200 to 700 sampels). The total amount of all collected paraphrases amounts to 253,500 paraphrases, reflecting the multiple data collection rounds datasets and LLMs we used. However, we did not investigate the effects of diversity incentives on larger amounts of collected data  to investigate if such data augmentation methods decrease in effectiveness when augmenting larger amounts of seed data.

    Eight, we only used the default settings of the diversity incentive methods and thus we did not compare the different number of seed sentences other than 6 per label and different number of hints to investigate the effectiveness of the diversity incentive methods under different settings.

    Ninth, the reproducibility of our data collection process for ChatGPT and GPT-4 is dependent upon the owners of ChatGPT services as the models we used in our study might be deprecated and not available in due time. This is, however, counterbalanced by the inclusion of 3 open LLMs.

    Tenth, we do not know if any of the 6 datasets used in this study have been used for training the LLMs we used for data collection and if this had any effect on our results and findings. As such, we do not know what kind of effect the diversity incentive methods would have on data augmentation of new, unpublished datasets. This limitation is part of the recently recognized possible ``LLM validation crisis'', as described by~\cite{li2023task}.

    Eleventh, we do not provide a direct comparison of LLMs against each other, as the seed sentences used for each data collection round changed for each LLM randomly. We believe, however, that this does not threaten the results, as the goal of this paper is to compare diversity incentive methods in a direct comparison on LLMs, not LLMs between each other.

\section*{Acknowledgements}
This work was partially supported by \textit{AI-CODE} (GA No. \href{https://cordis.europa.eu/project/id/101135437}{101135437}), \textit{VIGILANT} (GA No. \href{https://doi.org/10.3030/101073921}{101073921}), and \textit{vera.ai} (GA No. \href{https://doi.org/10.3030/101070093}{101070093}), projects funded by the European Union under the Horizon Europe. This work was also partially supported by \textit{MODERMED} a project funded by the Slovak Research and Development Agency, GA No. APVV-22-0414.

Part of the research results were obtained using the computational resources procured in the national project National competence centre for high performance computing (project code: 311070AKF2) funded by European Regional Development Fund, EU Structural Funds Informatization of society, Operational Program Integrated Infrastructure. This work was supported by the Ministry of Education, Youth and Sports of the Czech Republic through the e-INFRA CZ (ID:90254).

\bibliography{anthology,custom}

\begin{thebibliography}{40}
\expandafter\ifx\csname natexlab\endcsname\relax\def\natexlab#1{#1}\fi

\bibitem[{Cegin et~al.(2023)Cegin, Simko, and Brusilovsky}]{cegin-etal-2023-chatgpt}
Jan Cegin, Jakub Simko, and Peter Brusilovsky. 2023.
\newblock \href {https://doi.org/10.18653/v1/2023.emnlp-main.117} {{C}hat{GPT} to replace crowdsourcing of paraphrases for intent classification: Higher diversity and comparable model robustness}.
\newblock In \emph{Proceedings of the 2023 Conference on Empirical Methods in Natural Language Processing}, pages 1889--1905, Singapore. Association for Computational Linguistics.

\bibitem[{Chang and Jia(2023)}]{chang-jia-2023-data}
Ting-Yun Chang and Robin Jia. 2023.
\newblock \href {https://doi.org/10.18653/v1/2023.acl-long.452} {Data curation alone can stabilize in-context learning}.
\newblock In \emph{Proceedings of the 61st Annual Meeting of the Association for Computational Linguistics (Volume 1: Long Papers)}, pages 8123--8144, Toronto, Canada. Association for Computational Linguistics.

\bibitem[{Chen et~al.(2020)Chen, Tian, Xiao, He, and Jin}]{chen-etal-2020-semantically}
Wenqing Chen, Jidong Tian, Liqiang Xiao, Hao He, and Yaohui Jin. 2020.
\newblock \href {https://doi.org/10.18653/v1/2020.coling-main.102} {A semantically consistent and syntactically variational encoder-decoder framework for paraphrase generation}.
\newblock In \emph{Proceedings of the 28th International Conference on Computational Linguistics}, pages 1186--1198, Barcelona, Spain (Online). International Committee on Computational Linguistics.

\bibitem[{Chowdhury et~al.(2022)Chowdhury, Zhuang, and Wang}]{chowdhury2022novelty}
Jishnu~Ray Chowdhury, Yong Zhuang, and Shuyi Wang. 2022.
\newblock Novelty controlled paraphrase generation with retrieval augmented conditional prompt tuning.
\newblock In \emph{Proceedings of the AAAI Conference on Artificial Intelligence}, volume~36, pages 10535--10544.

\bibitem[{Cox et~al.(2023)Cox, Abdul, and Ooi}]{cox-2023-prompting}
Samuel~Rhys Cox, Ashraf Abdul, and Wei~Tsang Ooi. 2023.
\newblock \href {https://doi.org/10.1145/3623809.3623931} {Prompting a large language model to generate diverse motivational messages: A comparison with human-written messages}.
\newblock In \emph{Proceedings of the 11th International Conference on Human-Agent Interaction}, HAI '23, page 378–380, New York, NY, USA. Association for Computing Machinery.

\bibitem[{Dai et~al.(2023)Dai, Liu, Liao, Huang, Cao, Wu, Zhao, Xu, Liu, Liu, Li, Zhu, Cai, Sun, Li, Shen, Liu, and Li}]{dai2023auggpt}
Haixing Dai, Zhengliang Liu, Wenxiong Liao, Xiaoke Huang, Yihan Cao, Zihao Wu, Lin Zhao, Shaochen Xu, Wei Liu, Ninghao Liu, Sheng Li, Dajiang Zhu, Hongmin Cai, Lichao Sun, Quanzheng Li, Dinggang Shen, Tianming Liu, and Xiang Li. 2023.
\newblock \href {http://arxiv.org/abs/2302.13007} {Auggpt: Leveraging chatgpt for text data augmentation}.

\bibitem[{Dettmers et~al.(2023)Dettmers, Pagnoni, Holtzman, and Zettlemoyer}]{dettmers2023qlora}
Tim Dettmers, Artidoro Pagnoni, Ari Holtzman, and Luke Zettlemoyer. 2023.
\newblock Qlora: Efficient finetuning of quantized llms.
\newblock \emph{arXiv preprint arXiv:2305.14314}.

\bibitem[{Fang et~al.(2023)Fang, Lee, and Zhai}]{fang2023using}
Luyang Fang, Gyeong-Geon Lee, and Xiaoming Zhai. 2023.
\newblock \href {http://arxiv.org/abs/2310.18365} {Using gpt-4 to augment unbalanced data for automatic scoring}.

\bibitem[{Gao et~al.(2021)Gao, Fisch, and Chen}]{gao-etal-2021-making}
Tianyu Gao, Adam Fisch, and Danqi Chen. 2021.
\newblock \href {https://doi.org/10.18653/v1/2021.acl-long.295} {Making pre-trained language models better few-shot learners}.
\newblock In \emph{Proceedings of the 59th Annual Meeting of the Association for Computational Linguistics and the 11th International Joint Conference on Natural Language Processing (Volume 1: Long Papers)}, pages 3816--3830, Online. Association for Computational Linguistics.

\bibitem[{Ghosh et~al.(2023)Ghosh, Evuru, Kumar, Ramaneswaran, Sakshi, Tyagi, and Manocha}]{ghosh-etal-2023-dale}
Sreyan Ghosh, Chandra~Kiran Evuru, Sonal Kumar, S~Ramaneswaran, S~Sakshi, Utkarsh Tyagi, and Dinesh Manocha. 2023.
\newblock Dale: Generative data augmentation for low-resource legal nlp.
\newblock In \emph{Proceedings of the 2023 Conference on Empirical Methods in Natural Language Processing}, Sentosa, Singapore.

\bibitem[{Goyal and Durrett(2020)}]{goyal-durrett-2020-neural}
Tanya Goyal and Greg Durrett. 2020.
\newblock \href {https://doi.org/10.18653/v1/2020.acl-main.22} {Neural syntactic preordering for controlled paraphrase generation}.
\newblock In \emph{Proceedings of the 58th Annual Meeting of the Association for Computational Linguistics}, pages 238--252, Online. Association for Computational Linguistics.

\bibitem[{Hemphill et~al.(1990)Hemphill, Godfrey, and Doddington}]{hemphill-etal-1990-atis}
Charles~T. Hemphill, John~J. Godfrey, and George~R. Doddington. 1990.
\newblock \href {https://aclanthology.org/H90-1021} {The {ATIS} spoken language systems pilot corpus}.
\newblock In \emph{Speech and Natural Language: Proceedings of a Workshop Held at Hidden Valley, {P}ennsylvania, June 24-27,1990}.

\bibitem[{Jiang et~al.(2023)Jiang, Sablayrolles, Mensch, Bamford, Chaplot, de~las Casas, Bressand, Lengyel, Lample, Saulnier, Lavaud, Lachaux, Stock, Scao, Lavril, Wang, Lacroix, and Sayed}]{jiang2023mistral}
Albert~Q. Jiang, Alexandre Sablayrolles, Arthur Mensch, Chris Bamford, Devendra~Singh Chaplot, Diego de~las Casas, Florian Bressand, Gianna Lengyel, Guillaume Lample, Lucile Saulnier, Lélio~Renard Lavaud, Marie-Anne Lachaux, Pierre Stock, Teven~Le Scao, Thibaut Lavril, Thomas Wang, Timothée Lacroix, and William~El Sayed. 2023.
\newblock \href {http://arxiv.org/abs/2310.06825} {Mistral 7b}.

\bibitem[{Joshi and He(2022)}]{joshi-he-2022-investigation}
Nitish Joshi and He~He. 2022.
\newblock \href {https://doi.org/10.18653/v1/2022.acl-long.256} {An investigation of the (in)effectiveness of counterfactually augmented data}.
\newblock In \emph{Proceedings of the 60th Annual Meeting of the Association for Computational Linguistics (Volume 1: Long Papers)}, pages 3668--3681, Dublin, Ireland. Association for Computational Linguistics.

\bibitem[{K{\"o}ksal et~al.(2023)K{\"o}ksal, Schick, and Schuetze}]{koksal-etal-2023-meal}
Abdullatif K{\"o}ksal, Timo Schick, and Hinrich Schuetze. 2023.
\newblock \href {https://doi.org/10.18653/v1/2023.findings-emnlp.36} {{MEAL}: Stable and active learning for few-shot prompting}.
\newblock In \emph{Findings of the Association for Computational Linguistics: EMNLP 2023}, pages 506--517, Singapore. Association for Computational Linguistics.

\bibitem[{Krishna et~al.(2020)Krishna, Wieting, and Iyyer}]{Krishna2020}
Kalpesh Krishna, John Wieting, and Mohit Iyyer. 2020.
\newblock \href {https://doi.org/10.18653/v1/2020.emnlp-main.55} {{Reformulating unsupervised style transfer as paraphrase generation}}.
\newblock \emph{EMNLP 2020 - 2020 Conference on Empirical Methods in Natural Language Processing, Proceedings of the Conference}, pages 737--762.

\bibitem[{Lacoste et~al.(2019)Lacoste, Luccioni, Schmidt, and Dandres}]{lacoste2019quantifying}
Alexandre Lacoste, Alexandra Luccioni, Victor Schmidt, and Thomas Dandres. 2019.
\newblock Quantifying the carbon emissions of machine learning.
\newblock \emph{arXiv preprint arXiv:1910.09700}.

\bibitem[{Lang(1995)}]{LANG1995331}
Ken Lang. 1995.
\newblock \href {https://doi.org/https://doi.org/10.1016/B978-1-55860-377-6.50048-7} {Newsweeder: Learning to filter netnews}.
\newblock In Armand Prieditis and Stuart Russell, editors, \emph{Machine Learning Proceedings 1995}, pages 331--339. Morgan Kaufmann, San Francisco (CA).

\bibitem[{Larson et~al.(2019)Larson, Mahendran, Lee, Kummerfeld, Hill, Laurenzano, Hauswald, Tang, and Mars}]{Larson2019}
Stefan Larson, Anish Mahendran, Andrew Lee, Jonathan~K. Kummerfeld, Parker Hill, Michael~A. Laurenzano, Johann Hauswald, Lingjia Tang, and Jason Mars. 2019.
\newblock \href {https://doi.org/10.18653/v1/n19-1051} {{Outlier detection for improved data quality and diversity in dialog systems}}.
\newblock \emph{NAACL HLT 2019 - 2019 Conference of the North American Chapter of the Association for Computational Linguistics: Human Language Technologies - Proceedings of the Conference}, 1:517--527.

\bibitem[{Larson et~al.(2020)Larson, Zheng, Mahendran, Tekriwal, Cheung, Guldan, Leach, and Kummerfeld}]{larson-etal-2020-iterative}
Stefan Larson, Anthony Zheng, Anish Mahendran, Rishi Tekriwal, Adrian Cheung, Eric Guldan, Kevin Leach, and Jonathan~K. Kummerfeld. 2020.
\newblock \href {https://doi.org/10.18653/v1/2020.emnlp-main.650} {Iterative feature mining for constraint-based data collection to increase data diversity and model robustness}.
\newblock In \emph{Proceedings of the 2020 Conference on Empirical Methods in Natural Language Processing (EMNLP)}, pages 8097--8106, Online. Association for Computational Linguistics.

\bibitem[{Lee et~al.(2023)Lee, Hunter, and Ruiz}]{platypus2023}
Ariel~N. Lee, Cole~J. Hunter, and Nataniel Ruiz. 2023.
\newblock Platypus: Quick, cheap, and powerful refinement of llms.

\bibitem[{Lewis et~al.(2020)Lewis, Liu, Goyal, Ghazvininejad, Mohamed, Levy, Stoyanov, and Zettlemoyer}]{lewis-etal-2020-bart}
Mike Lewis, Yinhan Liu, Naman Goyal, Marjan Ghazvininejad, Abdelrahman Mohamed, Omer Levy, Veselin Stoyanov, and Luke Zettlemoyer. 2020.
\newblock \href {https://doi.org/10.18653/v1/2020.acl-main.703} {{BART}: Denoising sequence-to-sequence pre-training for natural language generation, translation, and comprehension}.
\newblock In \emph{Proceedings of the 58th Annual Meeting of the Association for Computational Linguistics}, pages 7871--7880, Online. Association for Computational Linguistics.

\bibitem[{Li and Flanigan(2023)}]{li2023task}
Changmao Li and Jeffrey Flanigan. 2023.
\newblock \href {http://arxiv.org/abs/2312.16337} {Task contamination: Language models may not be few-shot anymore}.

\bibitem[{Li and Qiu(2023)}]{li-qiu-2023-finding}
Xiaonan Li and Xipeng Qiu. 2023.
\newblock \href {https://doi.org/10.18653/v1/2023.findings-emnlp.411} {Finding support examples for in-context learning}.
\newblock In \emph{Findings of the Association for Computational Linguistics: EMNLP 2023}, pages 6219--6235, Singapore. Association for Computational Linguistics.

\bibitem[{Mosbach et~al.(2020)Mosbach, Andriushchenko, and Klakow}]{mosbach2020stability}
Marius Mosbach, Maksym Andriushchenko, and Dietrich Klakow. 2020.
\newblock On the stability of fine-tuning bert: Misconceptions, explanations, and strong baselines.
\newblock In \emph{International Conference on Learning Representations}.

\bibitem[{Mosbach et~al.(2023)Mosbach, Pimentel, Ravfogel, Klakow, and Elazar}]{mosbach-etal-2023-shot}
Marius Mosbach, Tiago Pimentel, Shauli Ravfogel, Dietrich Klakow, and Yanai Elazar. 2023.
\newblock \href {https://doi.org/10.18653/v1/2023.findings-acl.779} {Few-shot fine-tuning vs. in-context learning: A fair comparison and evaluation}.
\newblock In \emph{Findings of the Association for Computational Linguistics: ACL 2023}, pages 12284--12314, Toronto, Canada. Association for Computational Linguistics.

\bibitem[{Pecher et~al.(2023)Pecher, Srba, and Bielikova}]{pecher2023effects}
Branislav Pecher, Ivan Srba, and Maria Bielikova. 2023.
\newblock On the effects of randomness on stability of learning with limited labelled data: A systematic literature review.
\newblock \emph{arXiv preprint arXiv:2312.01082}.

\bibitem[{Piedboeuf and Langlais(2023)}]{piedboeuf-langlais-2023-chatgpt}
Fr{\'e}d{\'e}ric Piedboeuf and Philippe Langlais. 2023.
\newblock \href {https://doi.org/10.18653/v1/2023.findings-emnlp.1044} {Is {C}hat{GPT} the ultimate data augmentation algorithm?}
\newblock In \emph{Findings of the Association for Computational Linguistics: EMNLP 2023}, pages 15606--15615, Singapore. Association for Computational Linguistics.

\bibitem[{Radford et~al.(2019)Radford, Wu, Child, Luan, Amodei, and Sutskever}]{radford2019language}
Alec Radford, Jeff Wu, Rewon Child, David Luan, Dario Amodei, and Ilya Sutskever. 2019.
\newblock Language models are unsupervised multitask learners.

\bibitem[{Rhys~Cox et~al.(2021)Rhys~Cox, Wang, Abdul, von~der Weth, and Y.~Lim}]{Cox2021}
Samuel Rhys~Cox, Yunlong Wang, Ashraf Abdul, Christian von~der Weth, and Brian Y.~Lim. 2021.
\newblock \href {https://doi.org/10.1145/3411764.3445782} {Directed diversity: Leveraging language embedding distances for collective creativity in crowd ideation}.
\newblock In \emph{Proceedings of the 2021 CHI Conference on Human Factors in Computing Systems}, CHI '21, New York, NY, USA. Association for Computing Machinery.

\bibitem[{Schuster et~al.(2019)Schuster, Gupta, Shah, and Lewis}]{schuster-etal-2019-cross-lingual}
Sebastian Schuster, Sonal Gupta, Rushin Shah, and Mike Lewis. 2019.
\newblock \href {https://doi.org/10.18653/v1/N19-1380} {Cross-lingual transfer learning for multilingual task oriented dialog}.
\newblock In \emph{Proceedings of the 2019 Conference of the North {A}merican Chapter of the Association for Computational Linguistics: Human Language Technologies, Volume 1 (Long and Short Papers)}, pages 3795--3805, Minneapolis, Minnesota. Association for Computational Linguistics.

\bibitem[{Socher et~al.(2013)Socher, Perelygin, Wu, Chuang, Manning, Ng, and Potts}]{socher-etal-2013-recursive}
Richard Socher, Alex Perelygin, Jean Wu, Jason Chuang, Christopher~D. Manning, Andrew Ng, and Christopher Potts. 2013.
\newblock \href {https://aclanthology.org/D13-1170} {Recursive deep models for semantic compositionality over a sentiment treebank}.
\newblock In \emph{Proceedings of the 2013 Conference on Empirical Methods in Natural Language Processing}, pages 1631--1642, Seattle, Washington, USA. Association for Computational Linguistics.

\bibitem[{Thompson and Post(2020)}]{thompson-post-2020-paraphrase}
Brian Thompson and Matt Post. 2020.
\newblock \href {https://aclanthology.org/2020.wmt-1.67} {Paraphrase generation as zero-shot multilingual translation: Disentangling semantic similarity from lexical and syntactic diversity}.
\newblock In \emph{Proceedings of the Fifth Conference on Machine Translation}, pages 561--570, Online. Association for Computational Linguistics.

\bibitem[{Touvron et~al.(2023)Touvron, Martin, Stone, Albert, Almahairi, Babaei, Bashlykov, Batra, Bhargava, Bhosale, Bikel, Blecher, Ferrer, Chen, Cucurull, Esiobu, Fernandes, Fu, Fu, Fuller, Gao, Goswami, Goyal, Hartshorn, Hosseini, Hou, Inan, Kardas, Kerkez, Khabsa, Kloumann, Korenev, Koura, Lachaux, Lavril, Lee, Liskovich, Lu, Mao, Martinet, Mihaylov, Mishra, Molybog, Nie, Poulton, Reizenstein, Rungta, Saladi, Schelten, Silva, Smith, Subramanian, Tan, Tang, Taylor, Williams, Kuan, Xu, Yan, Zarov, Zhang, Fan, Kambadur, Narang, Rodriguez, Stojnic, Edunov, and Scialom}]{touvron2023LLaMA2}
Hugo Touvron, Louis Martin, Kevin Stone, Peter Albert, Amjad Almahairi, Yasmine Babaei, Nikolay Bashlykov, Soumya Batra, Prajjwal Bhargava, Shruti Bhosale, Dan Bikel, Lukas Blecher, Cristian~Canton Ferrer, Moya Chen, Guillem Cucurull, David Esiobu, Jude Fernandes, Jeremy Fu, Wenyin Fu, Brian Fuller, Cynthia Gao, Vedanuj Goswami, Naman Goyal, Anthony Hartshorn, Saghar Hosseini, Rui Hou, Hakan Inan, Marcin Kardas, Viktor Kerkez, Madian Khabsa, Isabel Kloumann, Artem Korenev, Punit~Singh Koura, Marie-Anne Lachaux, Thibaut Lavril, Jenya Lee, Diana Liskovich, Yinghai Lu, Yuning Mao, Xavier Martinet, Todor Mihaylov, Pushkar Mishra, Igor Molybog, Yixin Nie, Andrew Poulton, Jeremy Reizenstein, Rashi Rungta, Kalyan Saladi, Alan Schelten, Ruan Silva, Eric~Michael Smith, Ranjan Subramanian, Xiaoqing~Ellen Tan, Binh Tang, Ross Taylor, Adina Williams, Jian~Xiang Kuan, Puxin Xu, Zheng Yan, Iliyan Zarov, Yuchen Zhang, Angela Fan, Melanie Kambadur, Sharan Narang, Aurelien Rodriguez, Robert Stojnic, Sergey Edunov, and Thomas
  Scialom. 2023.
\newblock \href {http://arxiv.org/abs/2307.09288} {Llama 2: Open foundation and fine-tuned chat models}.

\bibitem[{Tripto et~al.(2023)Tripto, Venkatraman, Macko, Moro, Srba, Uchendu, Le, and Lee}]{tripto2023ship}
Nafis~Irtiza Tripto, Saranya Venkatraman, Dominik Macko, Robert Moro, Ivan Srba, Adaku Uchendu, Thai Le, and Dongwon Lee. 2023.
\newblock \href {http://arxiv.org/abs/2311.08374} {A ship of theseus: Curious cases of paraphrasing in llm-generated texts}.

\bibitem[{Ubani et~al.(2023)Ubani, Polat, and Nielsen}]{ubani2023zeroshotdataaug}
Solomon Ubani, Suleyman~Olcay Polat, and Rodney Nielsen. 2023.
\newblock \href {http://arxiv.org/abs/2304.14334} {Zeroshotdataaug: Generating and augmenting training data with chatgpt}.

\bibitem[{Wang et~al.(2022)Wang, Huang, Zhang, Lee, and Xing}]{wang2022toward}
Haohan Wang, Zeyi Huang, Hanlin Zhang, Yong~Jae Lee, and Eric~P Xing. 2022.
\newblock Toward learning human-aligned cross-domain robust models by countering misaligned features.
\newblock In \emph{Uncertainty in Artificial Intelligence}, pages 2075--2084. PMLR.

\bibitem[{Wei et~al.(2018)Wei, Chi, Chien, Ming, Kun, Po, and Chu}]{Bohlen2018}
Lee Wei, Hsuan~Huang Chi, Wei~Chang Chien, Kuang Daniel~Wu Ming, Ta~Chuang Kun, An~Yang Po, and Cheng~Hsieh Chu. 2018.
\newblock \href {https://doi.org/10.5441/002/edbt.2018.75} {Effective quality assurance for data labels through crowdsourcing and domain expert collaboration}.
\newblock In \emph{Advances in Database Technology - EDBT 2018}, Advances in Database Technology - EDBT, pages 646--649. OpenProceedings.org.

\bibitem[{Zhang et~al.(2015)Zhang, Zhao, and LeCun}]{zhang2015character}
Xiang Zhang, Junbo Zhao, and Yann LeCun. 2015.
\newblock Character-level convolutional networks for text classification.
\newblock \emph{Advances in neural information processing systems}, 28.

\bibitem[{Zhou and Bhat(2020)}]{Yaghoub-Zadeh-Fard2020}
Jianing Zhou and Suma Bhat. 2020.
\newblock \href {https://doi.org/10.1145/3377325.3377486} {{Dynamic word recommendation to obtain diverse crowdsourced paraphrases of user utterances}}.
\newblock In \emph{International Conference on Intelligent User Interfaces, Proceedings IUI}, pages 55--66.

\end{thebibliography}

\appendix

\begin{table*}[t!]
    \begin{subtable}[h]{1\linewidth}
        \centering
	  \small
        \setlength\tabcolsep{3pt}
        \begin{tabular}{@{}lccccc@{}}
        \toprule
        \textit{20 News}        & \textsc{prompt}       & \textsc{taboo}        & \textsc{chaining}     & \textsc{hints} & \textsc{comb}\\ \midrule
        ChatGPT	& $60.01_{5.43}$	& $59.45_{5.54}$	& $57.78_{5.92}$	& $60.30_{4.96}$	& $59.44_{5.61}$\\
        GPT4	& $61.38_{2.80}$	& $65.44_{2.58}$	& $62.30_{3.84}$	& $65.43_{2.95}$	& $60.58_{3.85}$\\
        Mistral	& $58.91_{3.81}$	& $58.53_{3.19}$	& $57.77_{3.28}$	& $58.92_{2.90}$	& $58.54_{2.94}$\\
        LLaMA-2	& $60.62_{4.46}$	& $59.32_{4.55}$	& $59.58_{5.26}$	& $60.87_{3.88}$	& $60.16_{4.83}$\\
        Platypus	& $61.95_{3.36}$	& $61.08_{3.37}$	& $60.24_{3.62}$	& $61.02_{2.41}$	& $59.35_{3.10}$\\
        \bottomrule
        \end{tabular}
       \caption{Results on the 20 News dataset.}
       \label{tab:20-news-results}
    \end{subtable}
    \hfill
    \begin{subtable}[h]{1\linewidth}
        \centering
        \small
        \setlength\tabcolsep{3pt}
        \begin{tabular}{@{}lccccc@{}}
        \toprule
        \textit{AG News}        & \textsc{prompt}       & \textsc{taboo}        & \textsc{chaining}     & \textsc{hints} & \textsc{comb}\\ \midrule
        ChatGPT	& $79.45_{2.37}$	& $79.32_{2.46}$	& $78.14_{2.68}$	& $80.43_{2.36}$	& $78.74_{2.96}$\\
        GPT4	& $79.35_{3.09}$	& $79.53_{2.89}$	& $77.74_{2.95}$	& $79.36_{2.06}$	& $79.55_{2.40}$\\
        Mistral	& $83.38_{1.75}$	& $83.26_{1.83}$	& $82.79_{2.43}$	& $83.40_{1.62}$	& $79.54_{2.61}$\\
        LLaMA-2	& $81.08_{3.19}$	& $81.83_{3.23}$	& $81.21_{3.35}$	& $81.56_{3.21}$	& $79.65_{4.15}$\\
        Platypus	& $78.56_{4.34}$	& $79.82_{3.45}$	& $78.65_{4.35}$	& $79.56_{3.80}$	& $78.57_{3.88}$\\
        \bottomrule
        \end{tabular}
       \caption{Results on the AG News dataset.}
       \label{tab:ag-news-results}
     \hfill
     \end{subtable}
         \begin{subtable}[h]{1\linewidth}
        \centering
        \small
        \setlength\tabcolsep{3pt}
        \begin{tabular}{@{}lccccc@{}}
        \toprule
        \textit{ATIS}   & \textsc{prompt}       & \textsc{taboo}        & \textsc{chaining}     & \textsc{hints} & \textsc{comb}\\ \midrule
        ChatGPT	& $82.94_{11.06}$	& $76.47_{13.05}$	& $80.94_{12.84}$	& $85.21_{7.73}$	& $79.44_{12.85}$\\
        GPT4	& $76.06_{8.78}$	& $74.61_{7.90}$	& $74.30_{9.80}$	& $76.47_{8.76}$	& $74.13_{7.99}$\\
        Mistral	& $79.83_{9.75}$	& $74.58_{7.60}$	& $75.10_{11.15}$	& $80.32_{9.21}$	& $75.31_{8.43}$\\
        LLaMA-2	& $82.88_{5.36}$	& $78.20_{6.31}$	& $81.11_{6.03}$	& $83.35_{5.53}$	& $79.34_{5.73}$\\
        Platypus	& $83.53_{8.30}$	& $81.29_{8.47}$	& $81.18_{8.98}$	& $83.73_{6.81}$	& $82.05_{8.40}$\\
        \bottomrule
        \end{tabular}
       \caption{Results on the ATIS dataset.}
       \label{tab:atis-results}
     \end{subtable}
     \hfill
     \begin{subtable}[h]{1\linewidth}
        \centering
        \small
        \setlength\tabcolsep{3pt}
        \begin{tabular}{@{}lccccc@{}}
        \toprule
        \textit{FB}     & \textsc{prompt}       & \textsc{taboo}        & \textsc{chaining}     & \textsc{hints} & \textsc{comb}\\ \midrule
        ChatGPT	& $83.10_{2.32}$	& $81.70_{2.06}$	& $82.25_{2.19}$	& $83.11_{1.51}$	& $80.74_{1.54}$\\
        GPT4	& $82.56_{2.92}$	& $80.55_{4.40}$	& $80.80_{3.37}$	& $82.51_{2.47}$	& $81.14_{3.15}$\\
        Mistral	& $79.18_{3.12}$	& $77.98_{4.14}$	& $78.72_{4.10}$	& $79.44_{3.68}$	& $77.67_{2.81}$\\
        LLaMA-2	& $79.60_{4.04}$	& $79.34_{4.12}$	& $79.44_{2.67}$	& $80.58_{2.67}$	& $78.75_{3.69}$\\
        Platypus	& $80.75_{2.10}$	& $79.60_{2.79}$	& $79.86_{4.74}$	& $82.23_{2.25}$	& $79.94_{2.16}$\\
        \bottomrule
        \end{tabular}
       \caption{Results on the FB dataset.}
       \label{tab:fb-results}
     \end{subtable}
     \hfill
     \begin{subtable}[h]{1\linewidth}
        \centering
        \small
        \setlength\tabcolsep{3pt}
        \begin{tabular}{@{}lccccc@{}}
        \toprule
        \textit{SST-5}  & \textsc{prompt}       & \textsc{taboo}        & \textsc{chaining}     & \textsc{hints} & \textsc{comb}\\ \midrule
        ChatGPT	& $34.94_{2.51}$	& $36.06_{2.70}$	& $35.28_{2.03}$	& $35.85_{2.06}$	& $34.32_{2.08}$\\
        GPT4	& $33.70_{2.09}$	& $34.36_{1.96}$	& $33.93_{1.97}$	& $33.88_{1.77}$	& $33.88_{1.47}$\\
        Mistral	& $33.19_{2.94}$	& $32.74_{2.94}$	& $32.54_{2.98}$	& $33.46_{2.43}$	& $32.74_{2.98}$\\
        LLaMA-2	& $33.37_{2.74}$	& $34.83_{2.17}$	& $32.97_{2.49}$	& $33.63_{2.15}$	& $33.79_{2.43}$\\
        Platypus	& $33.89_{2.69}$	& $33.92_{2.13}$	& $33.41_{2.54}$	& $34.24_{2.23}$	& $34.11_{2.13}$\\
        \bottomrule
        \end{tabular}
       \caption{Results on the SST-5 dataset.}
       \label{tab:sst5-results}
     \end{subtable}
     \hfill
     \begin{subtable}[h]{1\linewidth}
        \centering
        \small
        \setlength\tabcolsep{3pt}
        \begin{tabular}{@{}lccccc@{}}
        \toprule
        \textit{Yelp}   & \textsc{prompt}       & \textsc{taboo}        & \textsc{chaining}     & \textsc{hints} & \textsc{comb}\\ \midrule
        ChatGPT	& $43.96_{2.02}$	& $44.86_{2.01}$	& $43.81_{2.85}$	& $44.17_{1.99}$	& $43.83_{1.70}$\\
        GPT4	& $41.50_{2.66}$	& $41.59_{2.47}$	& $40.69_{3.35}$	& $41.94_{2.83}$	& $41.34_{2.08}$\\
        Mistral	& $39.43_{2.83}$	& $40.08_{2.76}$	& $38.94_{2.21}$	& $39.69_{2.62}$	& $39.47_{3.63}$\\
        LLaMA-2	& $43.00_{3.41}$	& $43.07_{2.82}$	& $42.09_{2.70}$	& $42.72_{2.82}$	& $42.27_{3.33}$\\
        Platypus	& $43.37_{3.05}$	& $42.79_{2.75}$	& $43.15_{2.79}$	& $43.35_{2.66}$	& $41.71_{3.91}$\\
        \bottomrule
        \end{tabular}
       \caption{Results on the Yelp dataset.}
       \label{tab:yelp-results}
     \end{subtable}
     \caption{Performance of BERT-large classifier on the test split of each dataset after being trained 5 times for each of the repeated 5 data collection rounds. We report the mean performance and standard deviation. The \textit{hints} method generally increases mean performance and stability of performance when compared to baseline \textit{prompt} method.}
     \label{tab:full_res_perf_bert}
\end{table*}

\begin{table*}[t!]
    \begin{subtable}[h]{1\linewidth}
        \centering
        \small
        \setlength\tabcolsep{3pt}
        \begin{tabular}{@{}lccccc@{}}
        \toprule
        \texttt{20 News}	& \textsc{prompt}	& \textsc{taboo}	& \textsc{chaining}	& \textsc{hints}& \textsc{comb}\\ \midrule
        ChatGPT	& $74.05_{4.13}$	& $74.02_{4.75}$	& $75.05_{3.25}$	& $76.05_{1.72}$	& $73.30_{2.03}$\\
        GPT4	& $75.25_{3.70}$	& $75.42_{2.49}$	& $73.38_{2.43}$	& $77.52_{2.74}$	& $73.41_{8.65}$\\
        Mistral	& $72.87_{0.61}$	& $74.20_{0.29}$	& $73.33_{0.77}$	& $73.89_{0.60}$	& $73.82_{1.19}$\\
        LLaMA-2	& $75.21_{2.37}$	& $75.64_{5.18}$	& $71.21_{6.69}$	& $77.19_{1.81}$	& $76.18_{2.64}$\\
        Platypus	& $73.43_{3.48}$	& $76.90_{2.93}$	& $72.43_{5.71}$	& $77.45_{2.21}$	& $76.30_{3.31}$\\
        \bottomrule
        \end{tabular}
        \caption{Results for 20 News dataset.}
        \label{tab:20-news-results_mist}
    \end{subtable}
    \hfill
    \begin{subtable}[h]{1\linewidth}
        \centering
        \small
        \setlength\tabcolsep{3pt}
        \begin{tabular}{@{}lccccc@{}}
        \toprule
        \texttt{AG News}	& \textsc{prompt}	& \textsc{taboo}	& \textsc{chaining}	& \textsc{hints} & \textsc{comb} \\ \midrule
        ChatGPT	& $82.09_{5.48}$	& $84.27_{3.16}$	& $82.31_{4.78}$	& $83.49_{3.75}$	& $82.34_{3.07}$\\
        GPT4	& $81.34_{4.88}$	& $81.39_{3.24}$	& $82.78_{5.73}$	& $84.95_{1.29}$	& $81.80_{4.00}$\\
        Mistral	& $85.96_{1.78}$	& $86.96_{0.93}$	& $85.71_{1.03}$	& $87.14_{0.76}$	& $85.96_{1.82}$\\
        LLaMA-2	& $81.64_{6.02}$	& $83.28_{3.41}$	& $85.09_{3.24}$	& $83.09_{1.29}$	& $84.55_{2.68}$\\
        Platypus	& $82.48_{2.38}$	& $85.27_{2.51}$	& $85.37_{2.05}$	& $85.23_{0.87}$	& $82.61_{3.91}$\\
        \bottomrule
        \end{tabular}
        \caption{Results for AG News dataset.}
        \label{tab:ag-news-results_mist}
     \hfill
     \end{subtable}
         \begin{subtable}[h]{1\linewidth}
        \centering
        \small
        \setlength\tabcolsep{3pt}
        \begin{tabular}{@{}lccccc@{}}
        \toprule
        \texttt{ATIS}	& \textsc{prompt}	& \textsc{taboo}	& \textsc{chaining}	& \textsc{hints} & \textsc{comb} \\ \midrule
        ChatGPT	& $89.91_{8.74}$	& $88.86_{9.67}$	& $91.05_{8.78}$	& $94.04_{3.15}$	& $89.21_{3.57}$\\
        GPT4	& $74.21_{9.06}$	& $77.37_{11.75}$	& $77.37_{10.83}$	& $82.89_{5.20}$	& $81.58_{6.96}$\\
        Mistral	& $89.56_{6.67}$	& $87.11_{7.69}$	& $87.89_{4.88}$	& $89.30_{4.85}$	& $85.79_{9.06}$\\
        LLaMA-2	& $82.89_{12.20}$	& $75.00_{10.98}$	& $86.32_{9.94}$	& $85.35_{5.01}$	& $81.58_{11.94}$\\
        Platypus	& $87.11_{9.47}$	& $85.00_{9.62}$	& $88.42_{6.02}$	& $89.91_{3.58}$	& $87.37_{6.58}$\\
        \bottomrule
        \end{tabular}
        \caption{Results for ATIS dataset.}
        \label{tab:atis-results_mist}
     \end{subtable}
     \hfill
     \begin{subtable}[h]{1\linewidth}
        \centering
        \small
        \setlength\tabcolsep{3pt}
        \begin{tabular}{@{}lccccc@{}}
        \toprule
        \texttt{FB}	& \textsc{prompt}	& \textsc{taboo}	& \textsc{chaining}	& \textsc{hints} & \textsc{comb}\\ \midrule
        ChatGPT	& $86.60_{2.90}$	& $87.02_{1.78}$	& $85.46_{3.00}$	& $87.97_{1.47}$	& $86.06_{3.84}$\\
        GPT4	& $87.93_{1.34}$	& $86.52_{4.74}$	& $83.65_{9.85}$	& $88.12_{2.20}$	& $85.00_{5.11}$\\
        Mistral	& $79.83_{5.58}$	& $79.18_{5.20}$	& $79.08_{4.28}$	& $80.52_{1.25}$	& $81.09_{3.16}$\\
        LLaMA-2	& $79.86_{1.49}$	& $82.34_{1.21}$	& $79.22_{3.08}$	& $84.08_{2.68}$	& $81.18_{4.87}$\\
        Platypus	& $83.81_{2.85}$	& $79.68_{4.76}$	& $81.74_{2.21}$	& $85.83_{2.30}$	& $84.54_{1.72}$\\
        \bottomrule
        \end{tabular}
        \caption{Results for FB dataset.}
        \label{tab:fb-results_mist}
     \end{subtable}
     \hfill
     \begin{subtable}[h]{1\linewidth}
       \centering
        \small
        \setlength\tabcolsep{3pt}
        \begin{tabular}{@{}lccccc@{}}
        \toprule
        \texttt{SST-5}	& \textsc{prompt}	& \textsc{taboo}	& \textsc{chaining}	& \textsc{hints} & \textsc{comb}\\ \midrule
        ChatGPT	& $49.92_{5.77}$	& $53.39_{2.77}$	& $49.95_{5.52}$	& $51.16_{1.83}$	& $51.83_{4.73}$\\
        GPT4	& $48.33_{4.81}$	& $53.03_{4.77}$	& $54.57_{1.91}$	& $51.79_{3.15}$	& $52.76_{4.18}$\\
        Mistral	& $48.62_{4.17}$	& $47.33_{2.19}$	& $51.86_{3.08}$	& $50.35_{3.81}$	& $46.85_{4.55}$\\
        LLaMA-2	& $47.72_{3.97}$	& $51.13_{2.39}$	& $48.42_{5.06}$	& $53.39_{3.00}$	& $51.31_{3.12}$\\
        Platypus	& $50.59_{5.92}$	& $51.86_{3.12}$	& $50.32_{2.60}$	& $50.90_{2.34}$	& $50.95_{3.60}$\\
        \bottomrule
        \end{tabular}
        \caption{Results for SST-5 dataset.}
        \label{tab:sst5-results_mist}
     \end{subtable}
     \hfill
     \begin{subtable}[h]{1\linewidth}
        \centering
        \small
        \setlength\tabcolsep{3pt}
        \begin{tabular}{@{}lccccc@{}}
        \toprule
        \texttt{Yelp}	& \textsc{prompt}	& \textsc{taboo}	& \textsc{chaining}	& \textsc{hints}& \textsc{comb}\\ \midrule
        ChatGPT	& $54.00_{2.42}$	& $52.52_{3.43}$	& $52.20_{3.34}$	& $54.01_{2.37}$	& $53.97_{2.86}$\\
        GPT4	& $53.71_{3.27}$	& $53.78_{3.97}$	& $53.01_{2.35}$	& $53.80_{1.85}$	& $53.73_{0.76}$\\
        Mistral	& $53.37_{2.10}$	& $54.16_{3.28}$	& $53.18_{3.02}$	& $55.28_{1.23}$	& $54.48_{1.11}$\\
        LLaMA-2	& $52.24_{4.42}$	& $53.32_{2.22}$	& $54.98_{2.33}$	& $55.40_{2.77}$	& $53.93_{3.16}$\\
        Platypus	& $54.59_{3.04}$	& $53.09_{3.26}$	& $53.47_{2.44}$	& $54.67_{1.52}$	& $54.01_{2.21}$\\
        \bottomrule
        \end{tabular}
        \caption{Results for Yelp dataset.}
        \label{tab:yelp-results_mist}
     \end{subtable}
     \caption{Performance of Mistral classifier on a subset of the test split of each dataset after being trained 5 times for each of the repeated 5 data collection rounds. We report the mean performance and standard deviation. The \textit{hints} method generally increases mean performance and stability of performance when compared to baseline \textit{prompt} method.}
     \label{tab:full_res_perf_Mistral}
\end{table*}

\section{Ethical considerations}
Based on a thorough ethical assessment, performed on the basis of intra-institutional ethical guidelines and checklists tailored to the use of data and algorithms, we see no ethical concerns pertaining directly to the conduct of this research. We also ethically assessed our paraphrase validity crowdsourcing process from Appendix~\ref{sec:appendix_para_valid} via our intra-institutional ethical guidelines and found no ethical concerns. In our study, we analyzed existing data or data generated using various LLMs. During our manual checking of the data in Section~\ref{sec:data_coll} we also ensured that the data contained no personal or offensive data. Albeit production of new data through LLMs bears several risks, such as introduction of biases, the small size of the produced dataset, sufficient for experimentation is, at the same time, insufficient for any major machine learning endeavors, where such biases could be transferred.

We follow the license terms for all the models and datasets we used (such as the one required for the use of the LLaMA-2 model) – all models and datasets allow their use as part of research.

\subsection{CO2 Emission Related to Experiments}

Data collection via open-source LLMs was conducted using a private infrastructure, which has a carbon efficiency of 0.432 kg CO$_2$/kWh. A cumulative of 100 hours of computation was performed on hardware of type A100 PCIe 40/80GB (TDP of 250W) for data collection.

Model finetuning for both BERT and Mistral was conducted using a private infrastructure, which has a carbon efficiency of 0.432 kg CO$_2$/kWh. A cumulative of 800 hours of computation was performed on hardware of type A100 PCIe 40/80GB (TDP of 250W) for data collection.

Total emissions together are estimated to be 97.2 kgCO$_2$ of which 0 percents were directly offset. We tried to reduce the generated emissions by using 4-bit quantization for LLMs and using a subset of test data for evaluation for Mistral finetuning, as inference is costly.

Estimations were conducted using the \href{https://mlco2.github.io/impact#compute}{MachineLearning Impact calculator} presented in \cite{lacoste2019quantifying}.

\section{Paraphrase validity checking process}\label{sec:appendix_para_valid}

For the process of checking the validity of the created paraphrases, we used our very own web app developed for this process. The users, who were the authors that also developed the app, were shown the seed samples and its label, from which LLM generated the paraphrases, and one particular paraphrase to validate. The authors/users all gave consent to the data collection process and had knowledge of how the data would be used. The instructions were \textit{"Please decide if the paraphrase has the same meaning as the seed sentence and if it adheres to the label of the seed sentence."}  The user was then able to either mark the paraphrase as valid or not, with an additional optional checkbox to label the paraphrase as ‘borderline case’ for possible re-visions. As the seed sentence changed only once in a while (we first showed all the paraphrases from one seed sentence) this significantly reduced the cognitive load on the annotator. The users/authors then discussed together the ‘borderline cases’ where the users were not sure about the validity of created paraphrases.

\section{Full results of model performance on }\label{sec:appendix_full_res}

In this section we report full result of our experiments for each dataset, diversity incentive method and LLM. The results for BERT-large are in Table~\ref{tab:full_res_perf_bert} and for Mistral in Table~\ref{tab:full_res_perf_Mistral}. Visualizations can be found in Appendix~\ref{sec:appendix_performance_vis_res}. The specific open-source LLMs used for data collection were LLaMA-2-70B-instruct~\footnote{https://huggingface.co/meta-llama/Llama-2-70b-chat-hf} with 70 bilion parameters,  Mistral-7B-instruct~\footnote{https://huggingface.co/mistralai/Mistral-7B-Instruct-v0.1} with 7 bilion parameters and  Platypus-70B-instruct~\footnote{https://huggingface.co/garage-bAInd/Platypus2-70B-instruct} with 70 bilion parameters.

\section{BERT-large finetuning details}\label{sec:appendix_bert_finetune}

We used the \textit{bert-large-uncased} version of the model from Huggingface and the best working hyperparameters from our hyperparameter search were batch size of 32, classifier dropout set to 0.2, used the AdamW optimizer with learning rate set to \textit{1e-5} and trained for 80 epochs. We evaluated the model during training after each 10 epochs and saved its performance. We reported the best test performance for each of the models during training.

\section{Mistral finetuning details}\label{sec:appendix_mistral_finetune}

We used the \textit{Mistral-7B-v0.1}~\footnote{https://huggingface.co/mistralai/Mistral-7B-v0.1} version of the model from Huggingface. For finetuning, we used the PEFT method QLoRA~\cite{dettmers2023qlora} in 4-bit setting with \textit{r=16} and \textit{alpha=16}. We finetuned the model for 20 epochs, used batch size of 32, learning rate of \textit{2e-5}, dropout of 0.1, used half-precision floating-point format (fp16), warmup ratio of 0.1, maximum grad. norm of 0.3, maximum sequence length of 128, weight decay set to 0.01 and used 8-bit Adam optimization. We evaluated the model on a subset of the test dataset (10\% of the original dataset) due to the lengthy inference times, similar to previous work~\cite{chang-jia-2023-data, li-qiu-2023-finding, gao-etal-2021-making, koksal-etal-2023-meal} on all of the datasets except for ATIS. We did, however, use different splits from the test part of the datasets to mitigate the effect of sample bias.

\section{Dataset details}\label{sec:appendix_dataset_details}

As we did not use all of the dataset labels and samples in each of the dataset, we list our setup here. We mostly used labels that were in the datasets with similar quantity to deal with the imbalanced datasets issue. All used datasets are in English language. For the \textit{20 News} dataset we used samples with labels \textit{politics}, \textit{wellness}, \textit{entertainment}, \textit{travel}, \textit{style and beauty} and \textit{parenting}. For the \textit{AG News}, \textit{SST-5} and \textit{Yelp} datasets we used all the samples. For the \textit{ATIS} dataset we used samples with labels \textit{atis\_abbreviation}, \textit{atis\_aircraft}, \textit{atis\_airfare}, \textit{atis\_airline}, \textit{atis\_flight}, \textit{atis\_flight\_time}, \textit{atis\_ground\_service} and \textit{atis\_quantity}. For the \textit{FB} dataset we used samples with labels \textit{get\_directions}, \textit{get\_distance}, \textit{get\_estimated\_arrival}, \textit{get\_estimated\_departure}, \textit{get\_estimated\_duration}, \textit{get\_info\_road\_condition}, \textit{get\_info\_route},
 \textit{get\_info\_traffic}, \textit{get\_location} and \textit{update\_directions}. 

\section{Number of collected samples per dataset}\label{sec:appendix_no_dataset_col}

For the \textit{20 News} datasets we used 36 seed samples (6 seed per label with 6 labels total) randomly sampled for each data collection round, resulting in 180 samples collected for each round and 396 samples in the final dataset (48 seed samples + 180 samples 1st round + 180 samples 2nd round). The entire test split we used for BERT finetuning had 11,751 samples.

For the \textit{AG News} dataset we used 24 seed samples randomly sampled for each data collection round, resulting in 120 samples collected for each round and 264 samples in the final dataset. The entire test split we used for BERT finetuning had 7,600 samples.

For the \textit{ATIS} dataset we used 48 seed samples randomly sampled for each data collection round, resulting in 240 samples collected for each round and 528 samples in the final dataset. The entire test split we used for BERT finetuning had 763 samples.

For the \textit{FB} dataset we used 60 seed samples randomly sampled for each data collection round, resulting in 300 samples collected for each round and 660 samples in the final dataset. The entire test split we used for BERT finetuning had 5,645 samples.

For the \textit{SST-5} dataset we used 30 seed samples randomly sampled for each data collection round, resulting in 150 samples collected for each round and 330 samples in the final dataset. The entire test split we used for BERT finetuning had 2,210 samples.

For the \textit{Yelp} dataset we used 30 seed samples randomly sampled for each data collection round, resulting in 150 samples collected for each round and 330 samples in the final dataset. The entire test split we used for BERT finetuning had 13,895 samples.

\newpage

\section{Templates and examples of diversity incentive prompts used in LLMs}\label{sec:appendix_example-incentives}
	
The \textit{prompt} and \textit{chaining} method: \textit{Rephrase an original question or statement 3 times. Original phrase: seed\_phrase.}

\begin{lstlisting}[frame=single,breaklines=true]
Prompt example

Rephrase an original question or statement 3 times. Original phrase: "tell me the fastest way to get home".
\end{lstlisting}

\begin{lstlisting}[frame=single,breaklines=true]
Chaining example

Rephrase an original question or statement 3 times. Original phrase: "please share the most rapid means of getting back to my dwelling".
\end{lstlisting}

The \textit{taboo} method: \textit{Rephrase an original question or statement 3 times. Original phrase: seed\_phrase. Don’t use the words “word\_1”, “word\_2" or “word\_3” in your responses.}

\begin{lstlisting}[frame=single,breaklines=true]
Taboo example

Rephrase an original question or statement 3 times. Original phrase: "tell me the fastest way to get home".

Don't use the words "arrive", "construction" or "house" in your responses.
\end{lstlisting}

The \textit{hints} method: \textit{Rephrase an original question or statement 3 times. Original phrase: seed\_phrase.
\#\#\#
Example paraphrases:
phrase\_1, phrase\_2, phrase\_3
\#\#\#}

\begin{lstlisting}[frame=single,breaklines=true]
Hints example

Rephrase an original question or statement 3 times. Original phrase: "tell me the fastest way to get home".

###
Example paraphrases:
"please share the most rapid means of getting back to my dwelling".
"inform me of the quickest route to reach my house".
"what is the swiftest method to arrive at my residence".
###
\end{lstlisting}

\section{Visualization of the effect of diversity incentive methods on lexical diversity }\label{sec:appendix_lex_vis_res}

\begin{figure*}
    \begin{tabular}{cc}
      \includegraphics[width=0.425\textwidth]{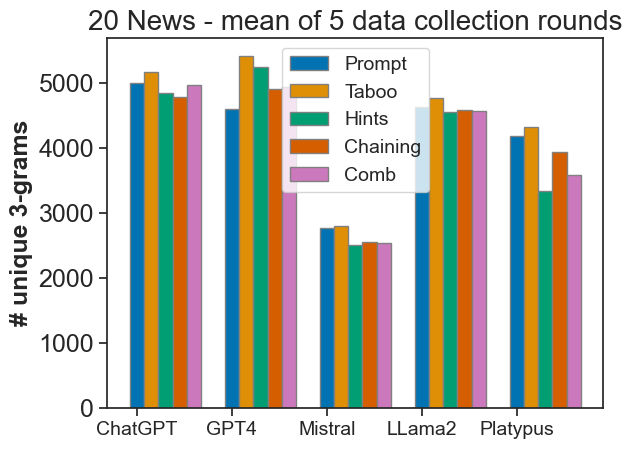} &   \includegraphics[width=0.425\textwidth]{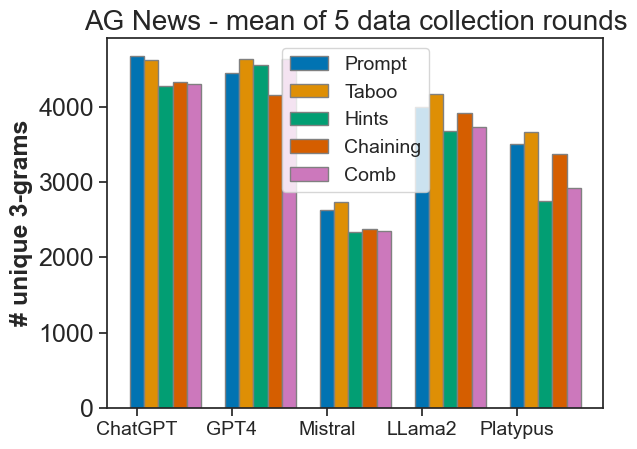} \\
    
     \includegraphics[width=0.425\textwidth]{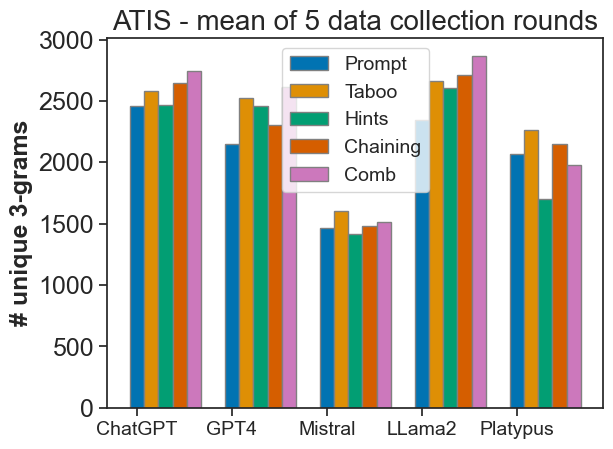} &   \includegraphics[width=0.425\textwidth]{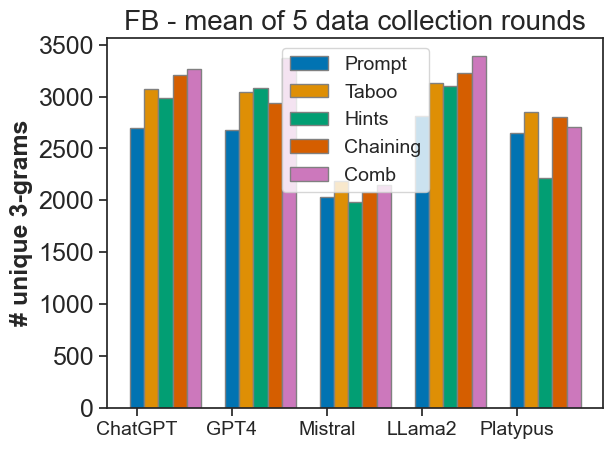} \\
   
     \includegraphics[width=0.425\textwidth]{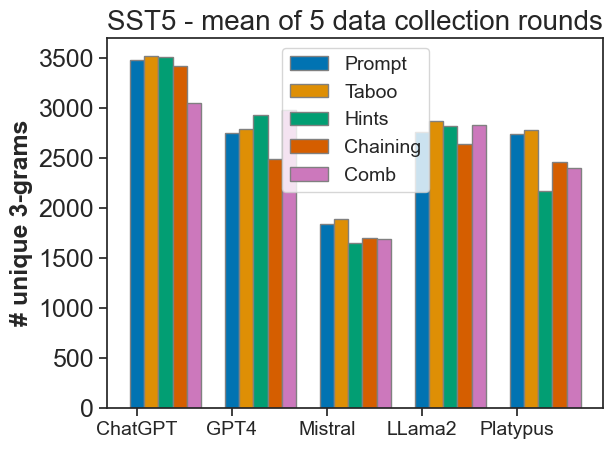} &   \includegraphics[width=0.425\textwidth]{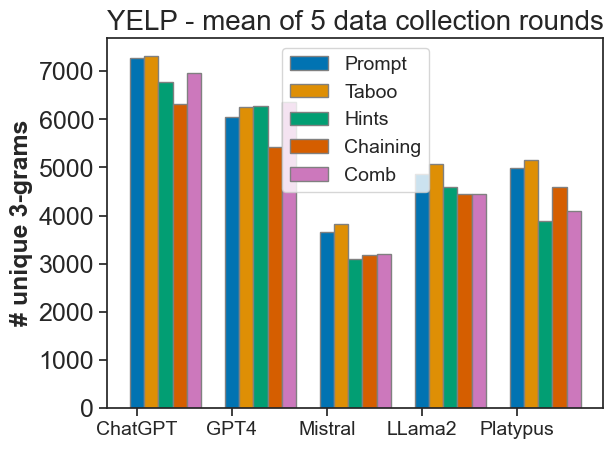} \\
    \end{tabular}
    \caption{Results of diversity incentive methods on no. of collected unique 3-grams per dataset and LLM combination. The \textit{taboo} method generally increases the no. of collected unique 3-grams, while the \textit{chaining} and \textit{hints} methods have random effects.}
    \label{fig:lexdiv}
    \end{figure*}

\begin{figure*}
    \begin{tabular}{cc}
      \includegraphics[width=0.425\textwidth]{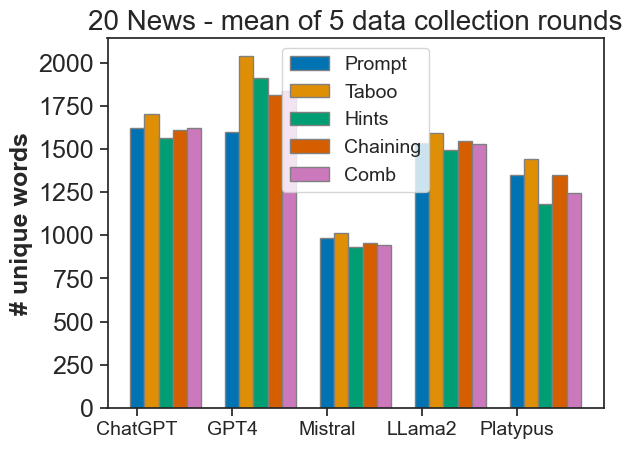} &   \includegraphics[width=0.425\textwidth]{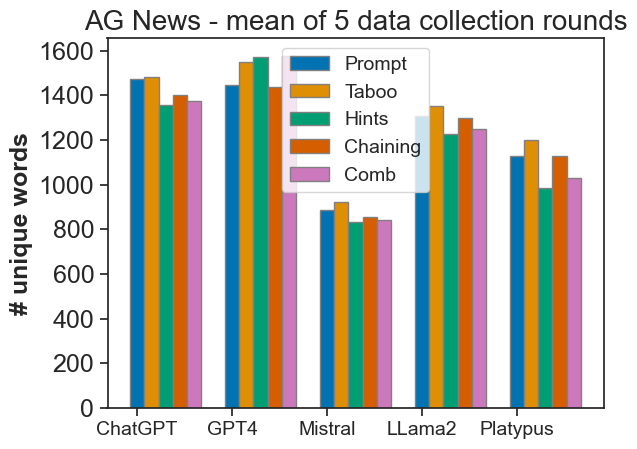} \\
     \includegraphics[width=0.425\textwidth]{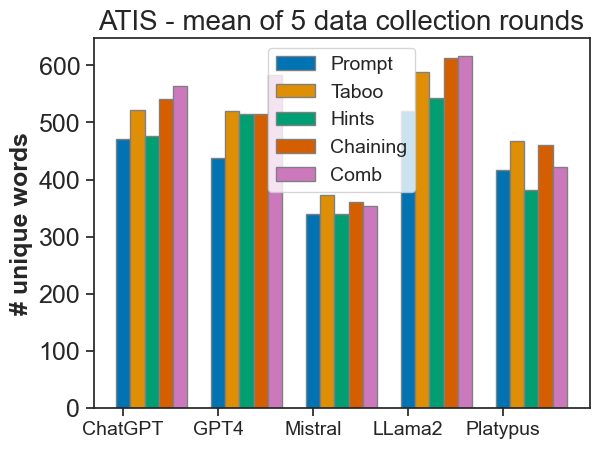} &   \includegraphics[width=0.425\textwidth]{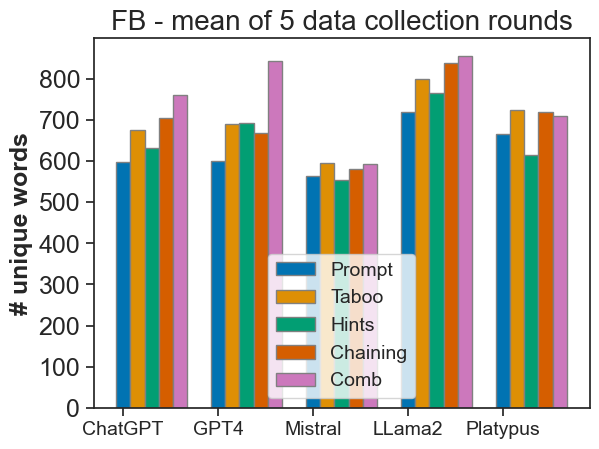} \\
     \includegraphics[width=0.425\textwidth]{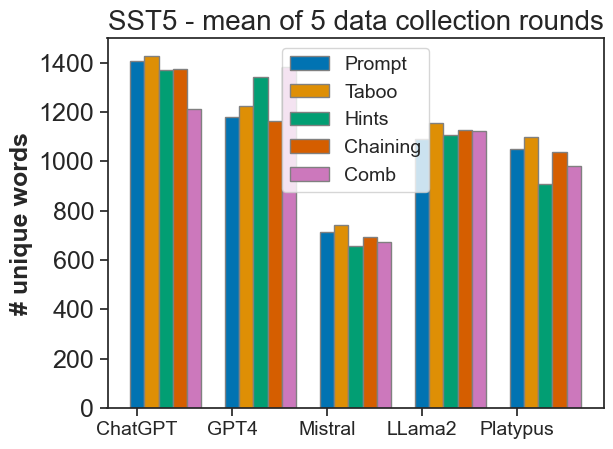} &   \includegraphics[width=0.425\textwidth]{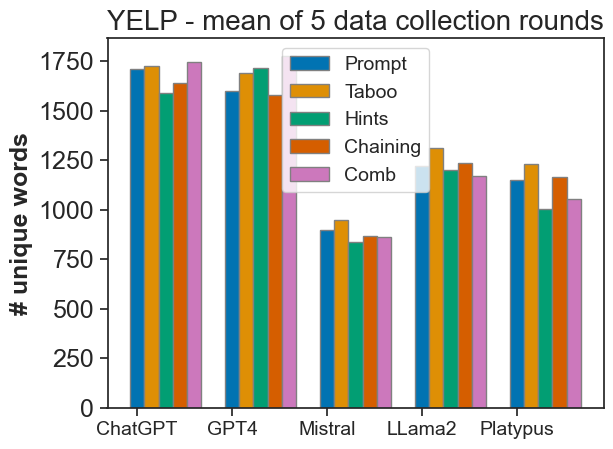} \\
    \end{tabular}\
    \caption{Results of diversity incentive methods on no. of collected unique words per dataset and LLM combination. The \textit{taboo} method generally increases the no. of collected unique words, while the \textit{chaining} and \textit{hints} methods have random effects.}
    \label{fig:worddiv}
    \end{figure*}

The effects of diversity incentive methods on lexical diversity can be found in Figure~\ref{fig:lexdiv} for no. unique 3-grams and Figure~\ref{fig:worddiv} for no. unique words.

\section{Visualization of the effect of diversity incentive methods model performance}\label{sec:appendix_performance_vis_res}

\begin{figure*}
    \begin{tabular}{cc}
      \includegraphics[width=0.425\textwidth]{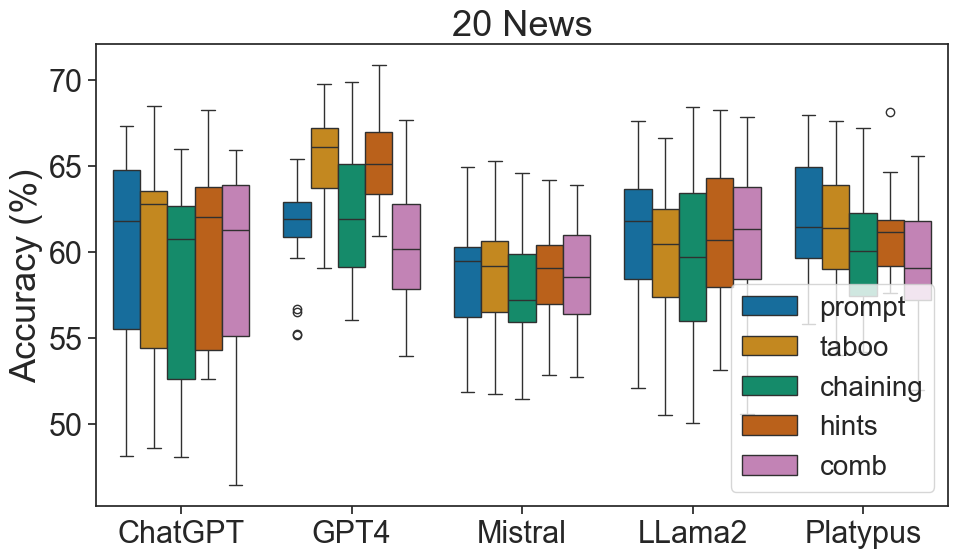} &   \includegraphics[width=0.425\textwidth]{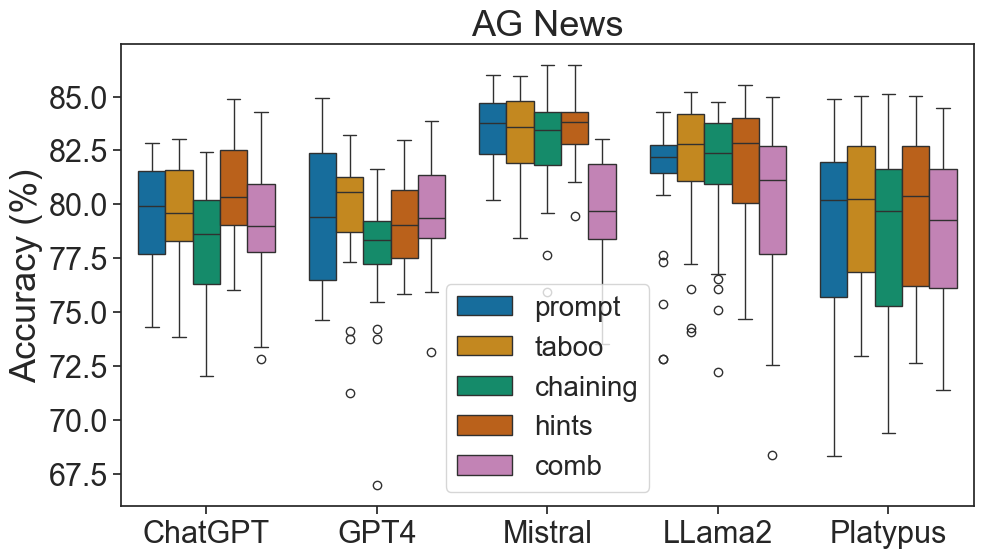} \\
     \includegraphics[width=0.425\textwidth]{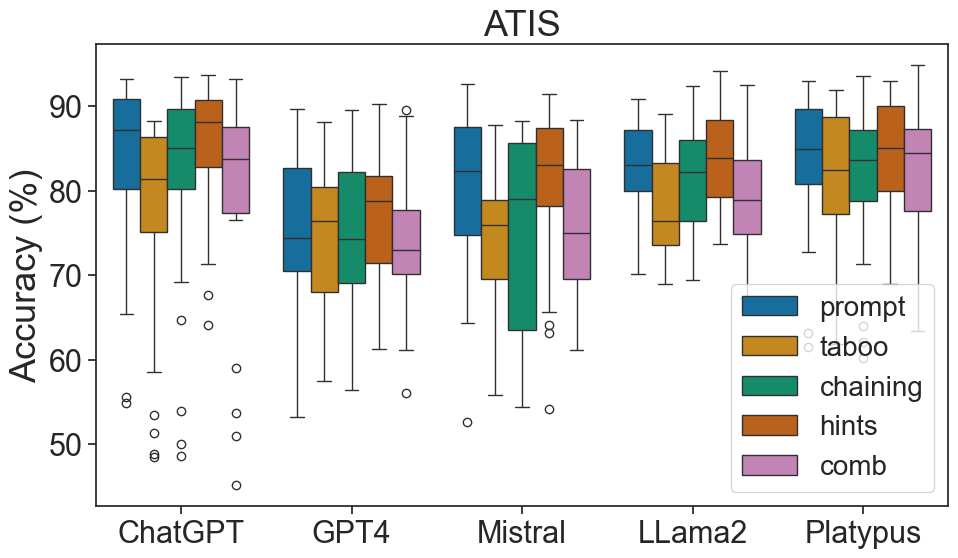} &   \includegraphics[width=0.425\textwidth]{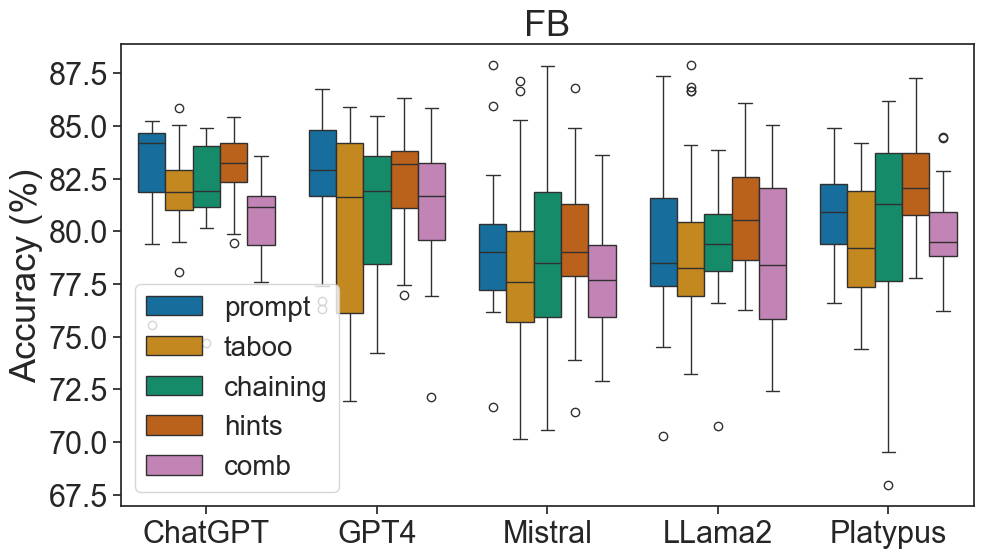} \\
     \includegraphics[width=0.425\textwidth]{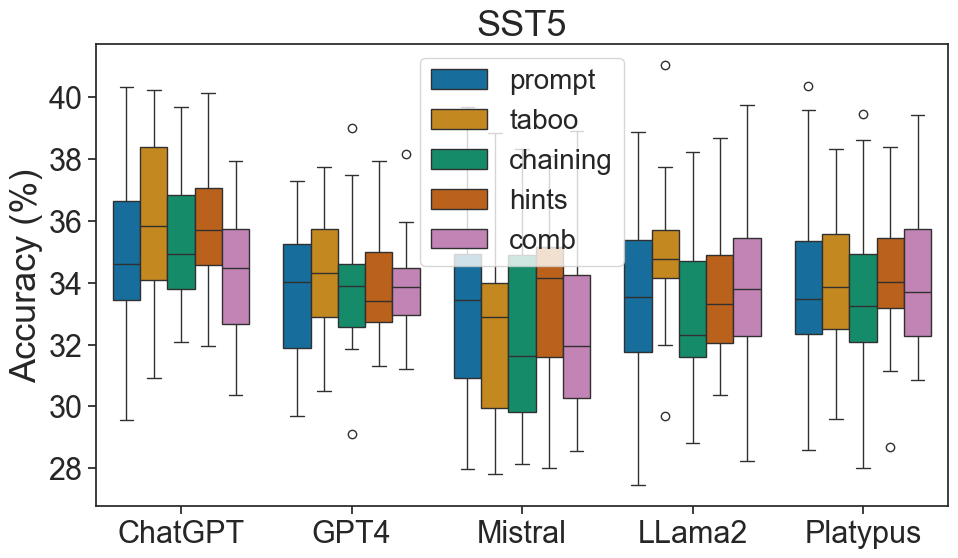} &   \includegraphics[width=0.425\textwidth]{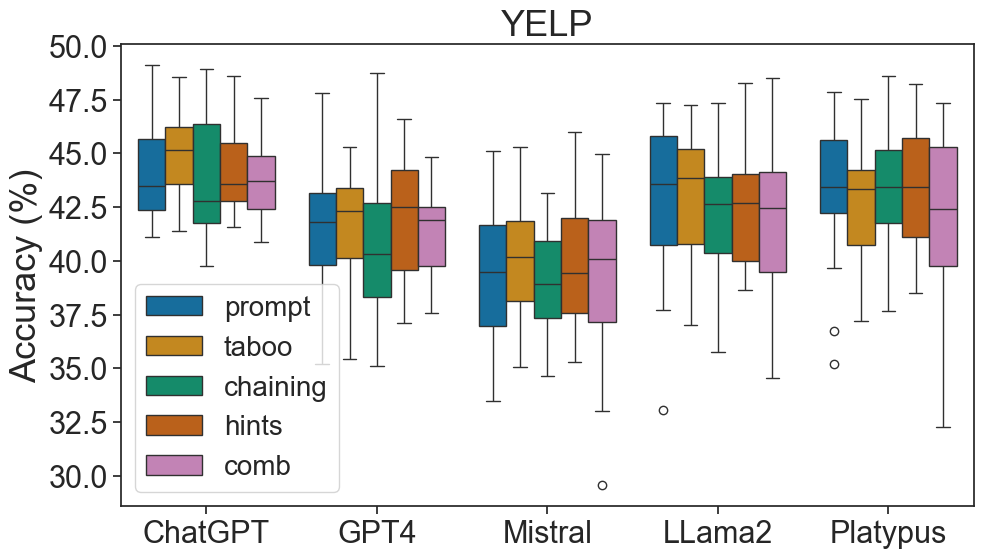} \\
    \end{tabular}
    \caption{The accuracy of BERT-large classifier on test data that was trained on data collected via different diversity incentive methods using various LLMs. The best performing methods is the \textit{hints} method, which generally increases mean performance of the models and stability of performance. The \textit{taboo} method has close to random influence on model performance while the \textit{chaining} method generally decreases model performance.}
    \label{fig:performance_boxplot_bert}
    \end{figure*}

\begin{figure*}
    \begin{tabular}{cc}
      \includegraphics[width=0.425\textwidth]{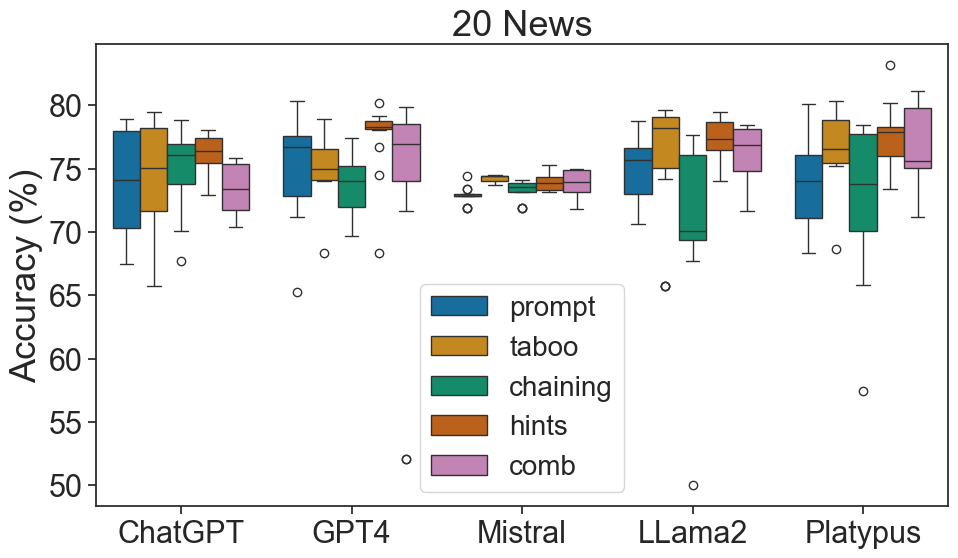} &   \includegraphics[width=0.425\textwidth]{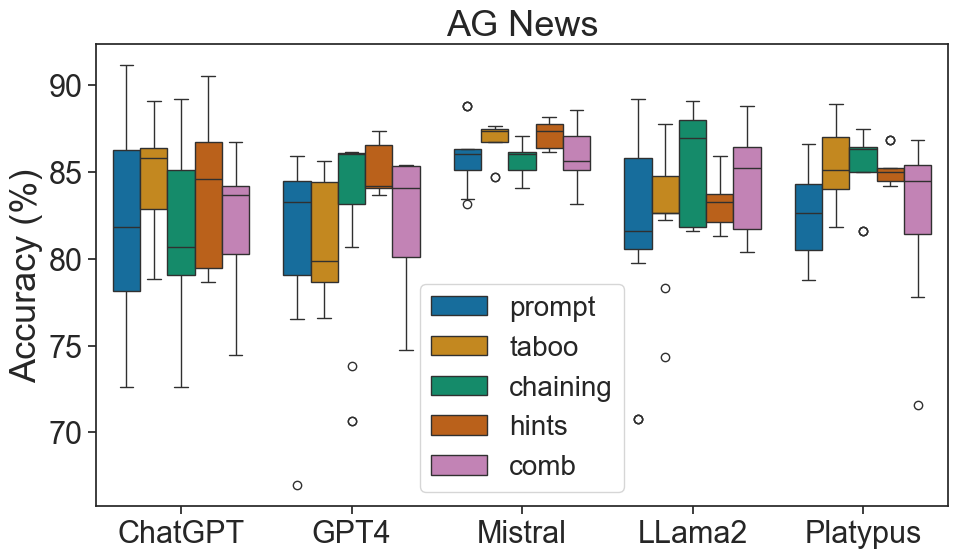} \\
     \includegraphics[width=0.425\textwidth]{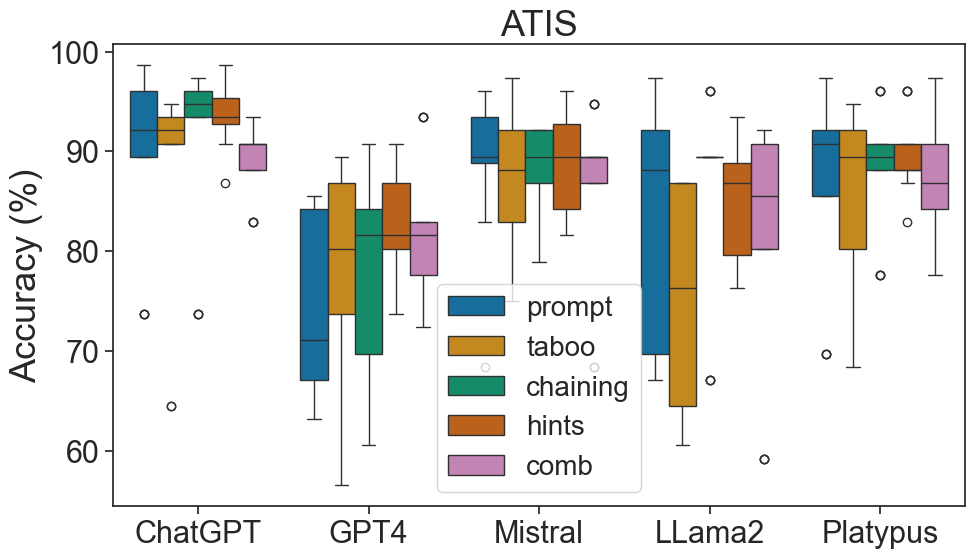} &   \includegraphics[width=0.425\textwidth]{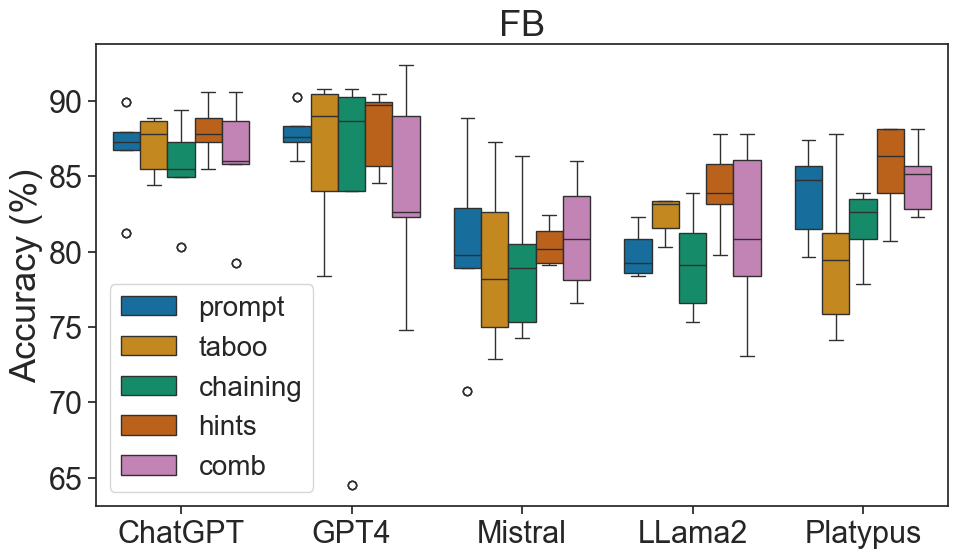} \\
     \includegraphics[width=0.425\textwidth]{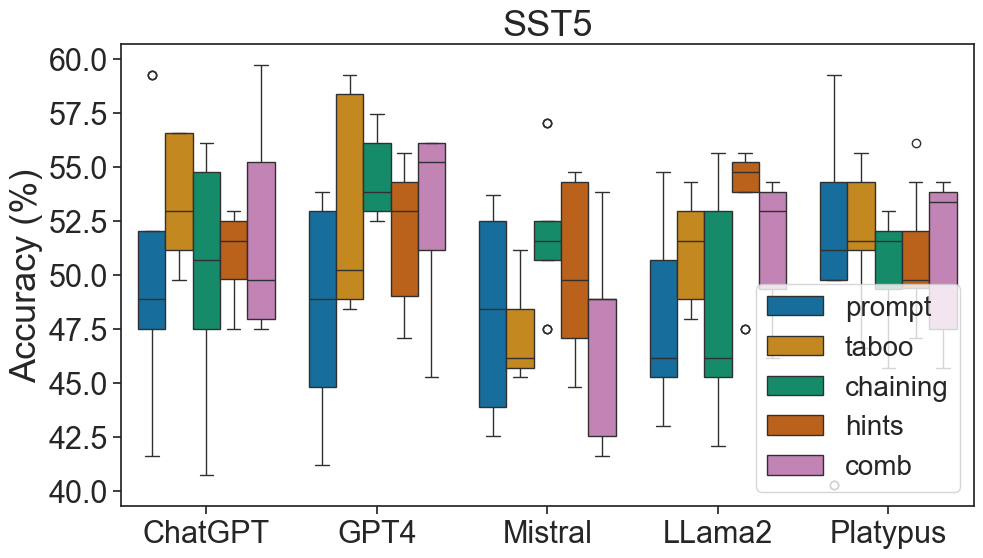} &   \includegraphics[width=0.425\textwidth]{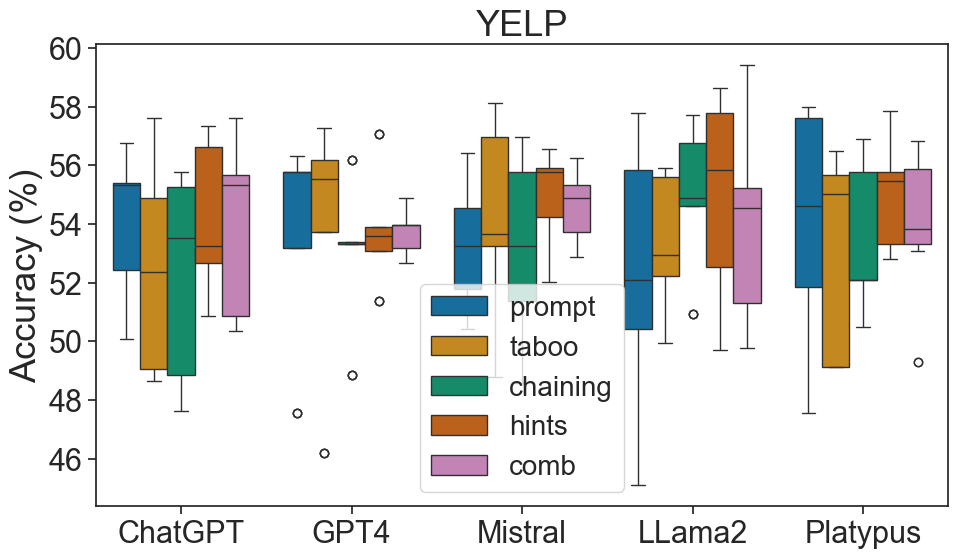} \\
    \end{tabular}
    \caption{The accuracy of Mistral finetuned for classification on a subset of test data. The model was trained on data collected via different diversity incentive methods using various LLMs. The best performing methods is the \textit{hints} method, which generally increases mean performance of the models and stability of performance.}
    \label{fig:performance_boxplot_mist}
    \end{figure*}

The effects of diversity incentive methods on model performance can be found in Figure~\ref{fig:performance_boxplot_bert} for BERT-large and Figure~\ref{fig:performance_boxplot_mist} for Mistral-7B.

\section{Results of the ablation study}
\label{sec:appendix_ablt_res_all}

\begin{figure*}
\begin{tabular}{cc}
  \includegraphics[width=0.425\textwidth]{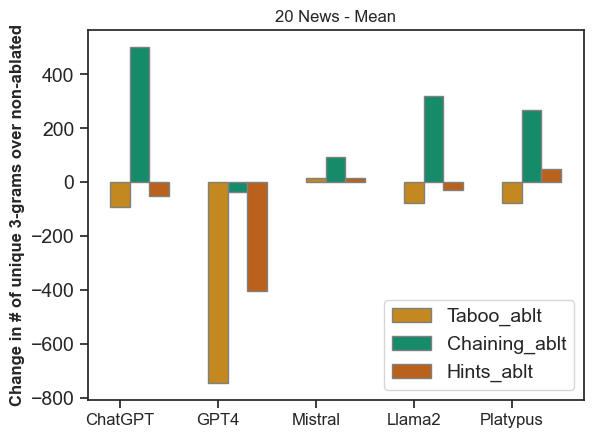} &   \includegraphics[width=0.425\textwidth]{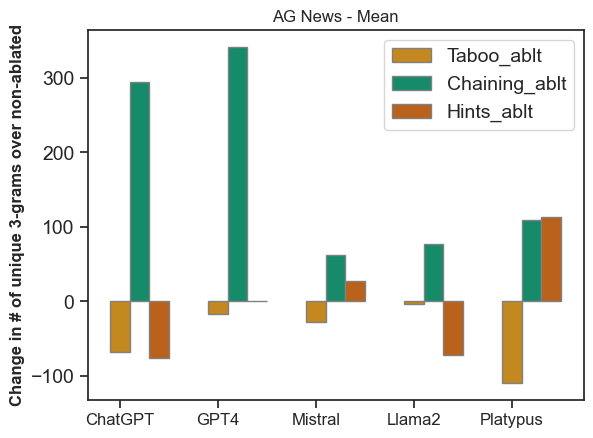} \\
 \includegraphics[width=0.425\textwidth]{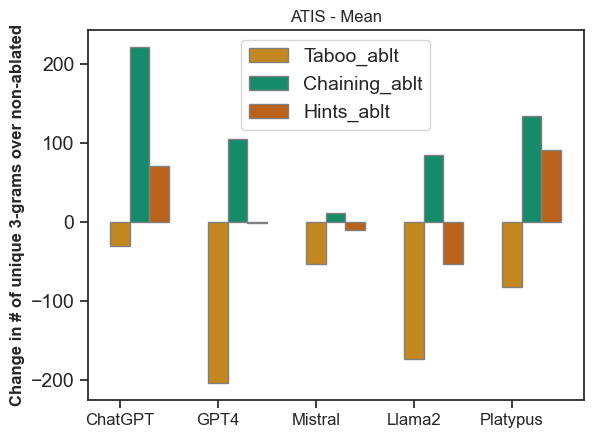} &   \includegraphics[width=0.425\textwidth]{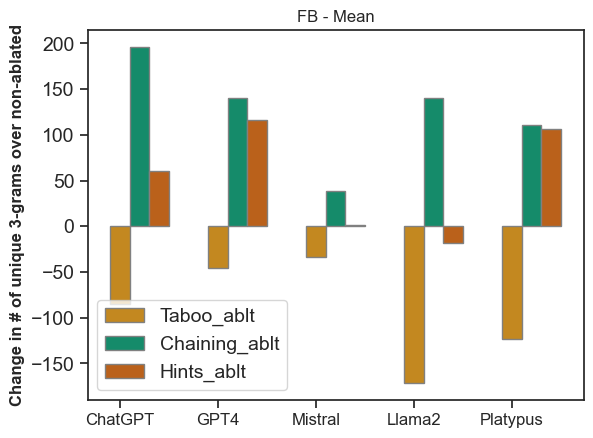} \\
 \includegraphics[width=0.425\textwidth]{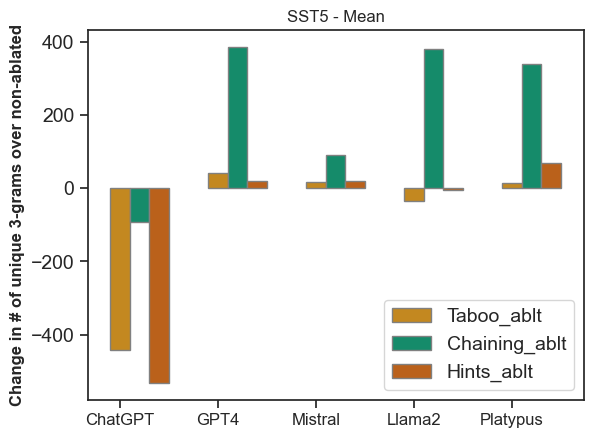} &   \includegraphics[width=0.425\textwidth]{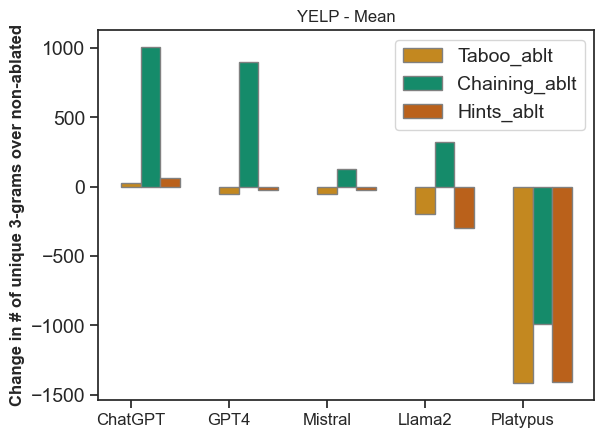} \\
\end{tabular}
\caption{The change in no. of collected unique 3-grams when comparing ablated methods with non-ablated. The figure displays the change of diversity of the ablated version of the diversity incentive methods vs. the non-ablated version. The ablated version of the \textit{taboo method} performs generally worse, indicating that the tabooing of most significant words increases diversity of texts collected via LLMs.}
\label{fig:ablation_res_ngrams_div}
\end{figure*}

\begin{figure*}
\begin{tabular}{cc}
  \includegraphics[width=0.425\textwidth]{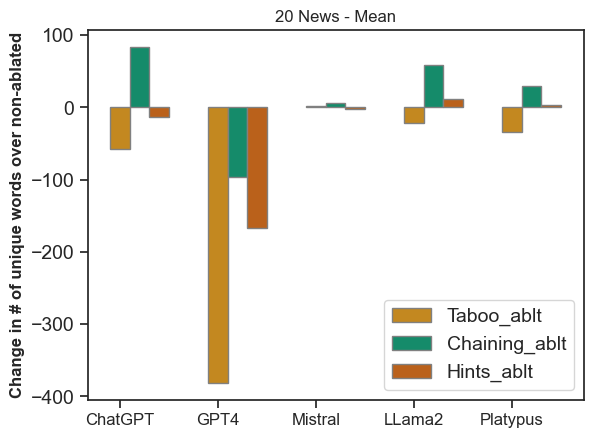} &   \includegraphics[width=0.425\textwidth]{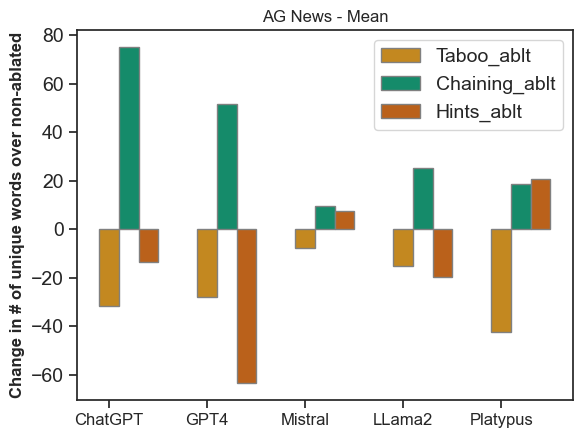} \\
 \includegraphics[width=0.425\textwidth]{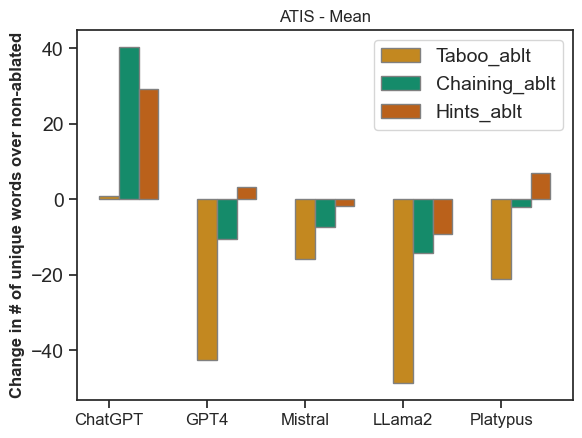} &   \includegraphics[width=0.425\textwidth]{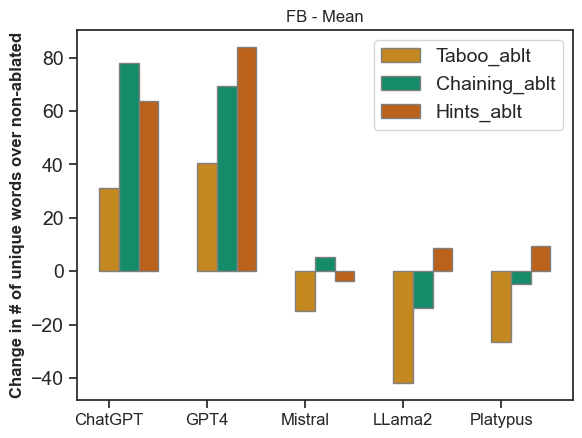} \\
 \includegraphics[width=0.425\textwidth]{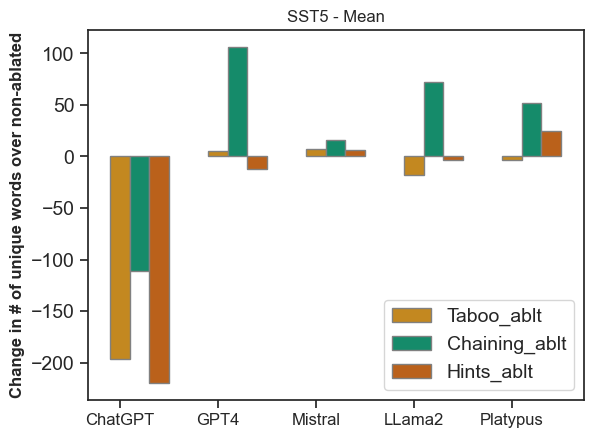} &   \includegraphics[width=0.425\textwidth]{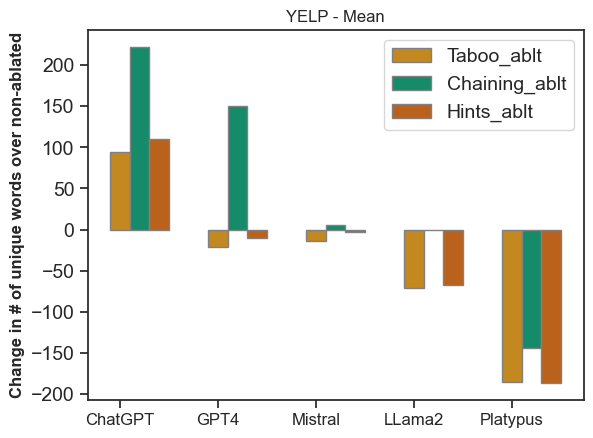} \\
\end{tabular}
\caption{The change in no. of collected unique words when comparing ablated methods with non-ablated. The figure displays the change of diversity of the ablated version of the diversity incentive methods vs. the non-ablated version. The ablated version of the \textit{taboo method} performs generally worse, indicating that the tabooing of most significant words increases diversity of texts collected via LLMs.}
\label{fig:ablation_res_words_div}
\end{figure*}

\begin{figure*}
\begin{tabular}{cc}
  \includegraphics[width=0.425\textwidth]{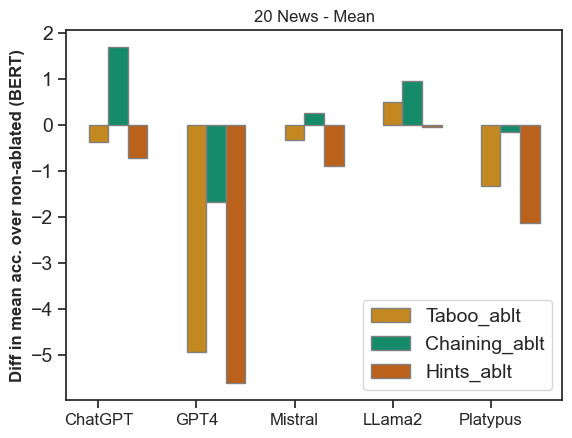} &   \includegraphics[width=0.425\textwidth]{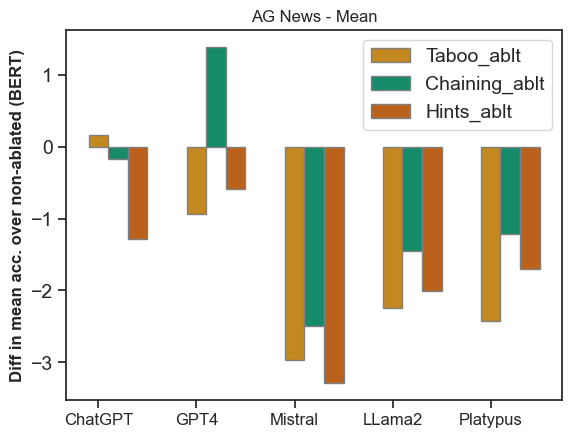} \\
 \includegraphics[width=0.425\textwidth]{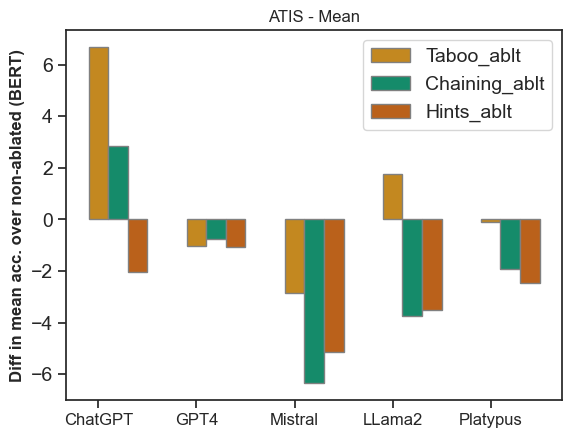} &   \includegraphics[width=0.425\textwidth]{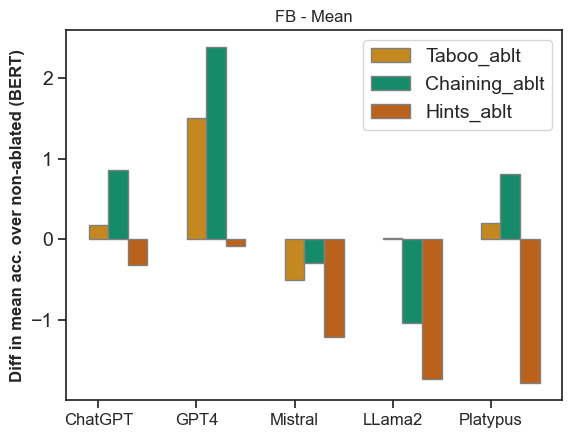} \\
 \includegraphics[width=0.425\textwidth]{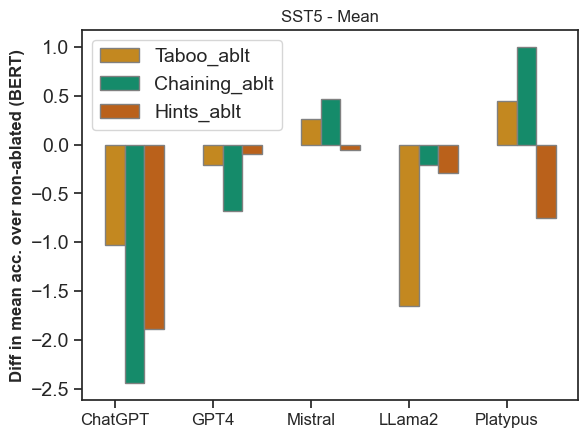} &   \includegraphics[width=0.425\textwidth]{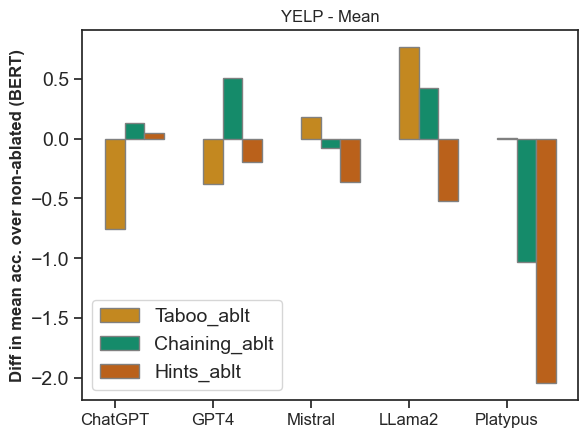} \\
\end{tabular}
\caption{The change in accuracy of BERT-large trained on data collected via ablated and non-ablated diversity incentive methods. The figure displays the change of accuracy of the ablated version of the diversity incentive methods vs. the non-ablated version. The ablated version of the \textit{hints method} performs generally worse, indicating that the inclusion of previous examples in the data collection yields better data.}
\label{fig:ablation_res_mean_acc}
\end{figure*}

\begin{figure*}
\begin{tabular}{cc}
  \includegraphics[width=0.425\textwidth]{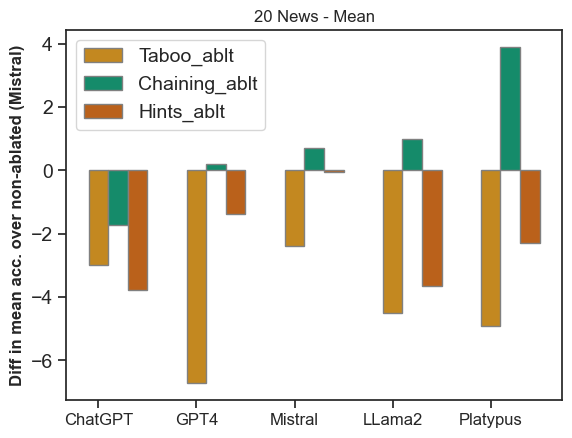} &   \includegraphics[width=0.425\textwidth]{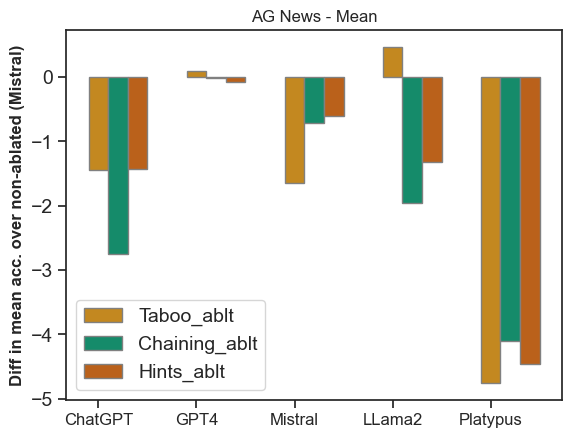} \\
 \includegraphics[width=0.425\textwidth]{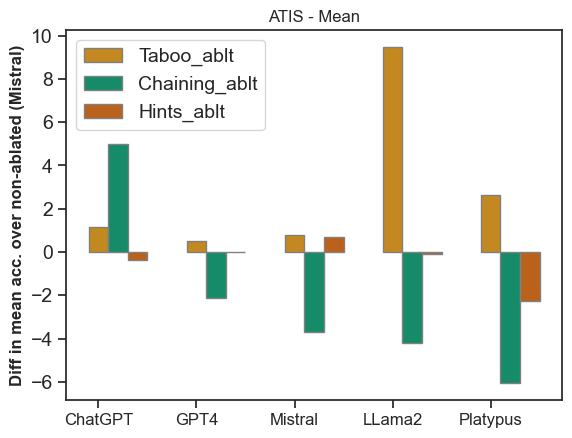} &   \includegraphics[width=0.425\textwidth]{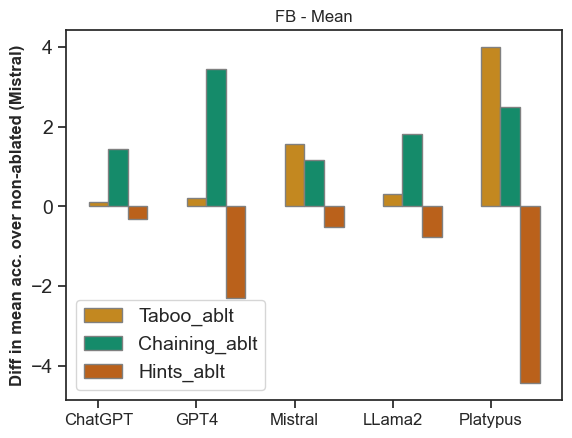} \\
 \includegraphics[width=0.425\textwidth]{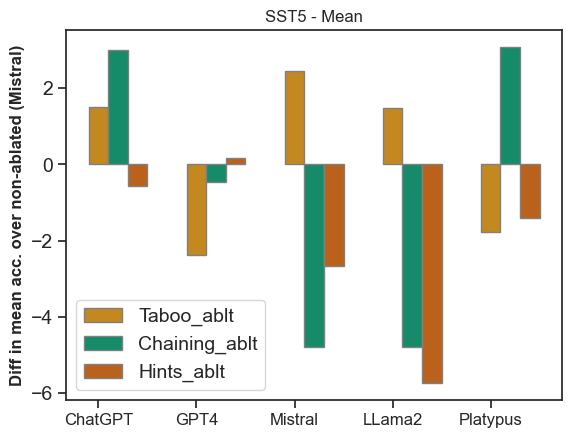} &   \includegraphics[width=0.425\textwidth]{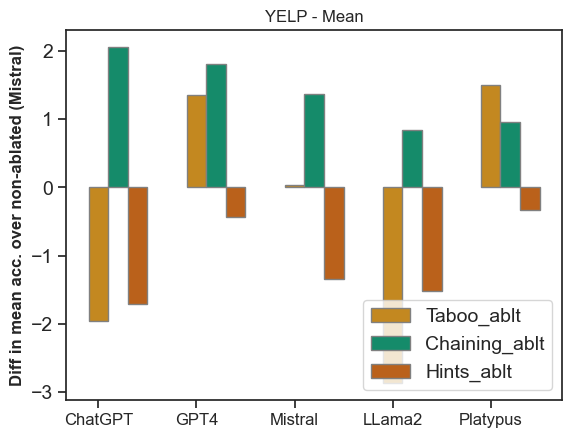} \\
\end{tabular}
\caption{The change in accuracy of Mistral trained on data collected via ablated and non-ablated diversity incentive methods. The figure displays the change of accuracy of the ablated version of the diversity incentive methods vs. the non-ablated version. The ablated version of the \textit{hints method} performs generally worse, indicating that the inclusion of previous examples in the data collection yields better data.}
\label{fig:ablation_res_mean_acc_mist}
\end{figure*}

In this section we list visualizations of the results for ablated versions of different diversity incentive methods in lexical diversity in Figures~\ref{fig:ablation_res_ngrams_div} and~\ref{fig:ablation_res_words_div} and in accuracy of models trained on data collected this way in Figures~\ref{fig:ablation_res_mean_acc} and~\ref{fig:ablation_res_mean_acc_mist}.

\end{document}